\documentclass[conference]{IEEEtran}
\IEEEoverridecommandlockouts
\usepackage{cite}

\usepackage{amsmath,amssymb,amsfonts}
\usepackage{graphicx}
\usepackage{textcomp}
\usepackage{xcolor}
\usepackage{xspace}
\usepackage{url}
\usepackage{tikz}
\usepackage{verbatim}
\usepackage[colorlinks=true, linkcolor=blue, citecolor=green, urlcolor=cyan]{hyperref}
\usepackage{filecontents}
\usepackage{tabularx}
\usepackage{booktabs}
\usepackage{enumitem}
\usepackage{multirow}
\usepackage{amssymb}
\usepackage[font=small]{caption}
\usepackage{soul}
\usepackage{pifont}
\usepackage{subcaption}
\usepackage[export]{adjustbox}
\usepackage{algorithm,algorithmicx}
\usepackage[noend]{algpseudocode}

\newcommand\crule[4][black]{\raisebox{#4}{\textcolor{#1}{\rule{#2}{#3}}}}

\definecolor{myviolet}{HTML}{D86ECC}
\definecolor{myred}{HTML}{FF0000}
\definecolor{mycyan}{HTML}{10FFFE}

\DeclareRobustCommand{\bdiamond}{%
  \mathbin{\text{\scalebox{1.2}{\rotatebox[origin=c]{45}{\textbf{\fboxsep=0pt\fboxrule=0.5pt\fbox{$\phantom{\rule{1ex}{1ex}}$}}}}}}%
}

\newcommand{\nofillbox}[5][black]{%
  \setlength{\fboxrule}{#2} 
  \definecolor{mycolor}{HTML}{#1}  
  \kern-0.35em 
  \raisebox{#5}{%
    \fcolorbox{mycolor}{white}{\phantom{\rule{#3}{#4}}}%
  }%
  \kern+0.10em 
}

\def\BibTeX{{\rm B\kern-.05em{\sc i\kern-.025em b}\kern-.08em
    T\kern-.1667em\lower.7ex\hbox{E}\kern-.125emX}}

\newcommand{\refer}{\emph{a.k.a.}\xspace}
\newcommand{\separate}{\textsc{Separate}\xspace}
\newcommand{\naivemix}{\textsc{NaiveMix}\xspace}

\newcommand{\utilmix}{\textsc{Mix-LUF}\xspace}
\newcommand{\approach}{\textsc{LeMix}\xspace}
\definecolor{royalblue}{HTML}{006795}

\begin{document}

\title{\approach: Unified Scheduling for LLM Training and Inference on Multi-GPU Systems}

\author{
Yufei Li
\xspace\xspace\xspace    
Zexin Li
\xspace\xspace\xspace   
Yinglun Zhu
\xspace\xspace\xspace  
Cong Liu
\\
University of California, Riverside
\\ 
\texttt{\{yli927,zli536,yzhu,congl\}@ucr.edu}
}

\maketitle

\begin{abstract}
Modern deployment of large language models (LLMs) frequently involves both inference serving and continuous retraining to stay aligned with evolving data and user feedback. Common practices separate these workloads onto distinct servers in isolated phases, causing substantial inefficiencies (e.g., GPU idleness) and delayed adaptation to new data in distributed settings. Our empirical analysis reveals that these inefficiencies stem from dynamic request arrivals during serving and workload heterogeneity in pipeline-parallel training. To address these challenges, we propose \approach, a system for \emph{co-locating} and managing concurrent LLM serving and training workloads. \approach integrates offline profiling, execution prediction mechanisms, and runtime scheduling to dynamically adapt resource allocation based on workload characteristics and system conditions. By understanding task-specific behaviors and co-execution interference across shared nodes, \approach improves utilization and serving quality without compromising serving responsiveness. Our evaluation shows that \approach improves throughput by up to 3.53$\times$, reduces inference loss by up to 0.61$\times$, and delivers up to 2.12$\times$ higher response time SLO attainment over traditional separate setups.
To our knowledge, this is the first work to uncover and exploit the opportunities of joint LLM inference and training, paving the way for more resource-efficient deployment of LLMs in production environments.
\end{abstract}


\section{Introduction}
Recent advances in large language models (LLMs) have underscored the importance of \emph{continuous} adaptation to maintain alignment with evolving data and user feedback~\cite{wei2022chain,yao2024tree,zheng2024judging}.
Techniques like test-time training~\cite{sun2020test,akyürek2024surprising} exemplify the potential for LLMs to improve dynamically during deployment. However, achieving this adaptability requires \emph{concurrent} training and inference workloads, which are traditionally partitioned and seperated across dedicated infrastructures~\cite{langton2018machine,choi2021multi,microsoft2023optimized}.
For instance, platforms such as Amazon SageMaker provide isolated environments for training and serving, where models are periodically retrained on new data and redeployed to inference endpoints~\cite{AmazonSageMakerTraining}. 
While this \separate setup simplifies management, it inherently results in resource inefficiency. 

\noindent\textbf{Potential resource inefficiencies.} 
In distributed systems, resource under-utilization arises from both independent training and inference workflows.
On the inference side, advanced serving systems~\cite{choi2022serving,kwon2023efficient,sun2024llumnix,zhong2024distserve,agrawal2024taming,wu2024dlora,fu2024serverlessllm} employ continuous batching~\cite{yu2022orca} to handle dynamic request arrivals.
Although these approaches improve throughput and response times under high traffic, they often assume sustained demand, leading to \emph{serving idleness} during off-peak periods when resources remain frequently inactive.
On the training side, parallelism techniques~\cite{gaunt2017ampnet,narayanan2019pipedream,shoeybi2019megatron,athlur2022varuna, choi2023envpipe,li2023alpaserve,lai2023merak} are widely used to accommodate the resource demands of LLMs by vertically partitioning computations across distributed systems.
While these approaches improve utilization, they introduce sequential dependencies in forward and backward passes, causing \emph{pipeline idleness} (\refer ``bubbles'')~\cite{huang2019gpipe} during execution.

\begin{figure*}[t]
    \centering
    \begin{subfigure}[t]{0.19\textwidth}
        \centering
        \includegraphics[height=90pt]{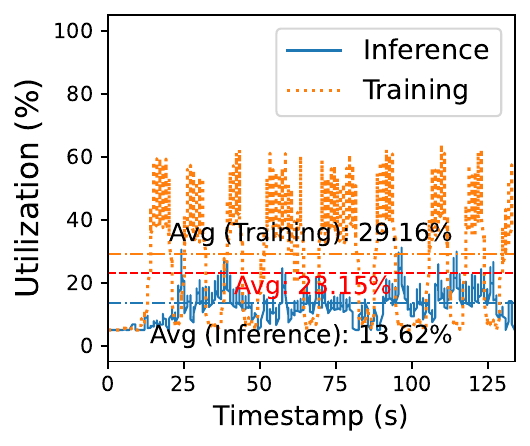}
        \caption{\separate (2-2)}
        \label{fig:sep_4_4}
    \end{subfigure}
    \hfill
    \begin{subfigure}[t]{0.19\textwidth}
        \centering
        \includegraphics[height=90pt]{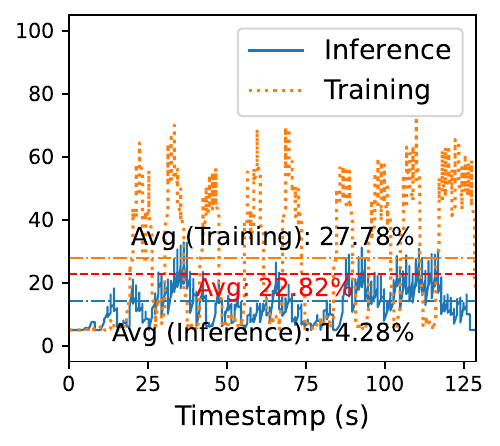}
        \caption{\separate (1-3)}
        \label{fig:sep_2_6}
    \end{subfigure}
    \hfill
    \begin{subfigure}[t]{0.19\textwidth}
        \centering
        \includegraphics[height=90pt]{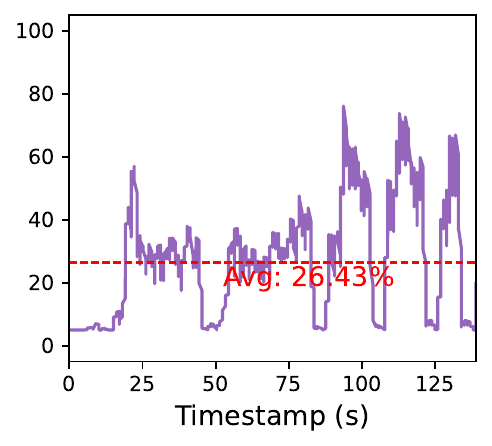}
        \caption{\separate (dynamic)}
        \label{fig:sep_dynamic}
    \end{subfigure}
    \hfill
    \begin{subfigure}[t]{0.19\textwidth}
        \centering
        \includegraphics[height=90pt]{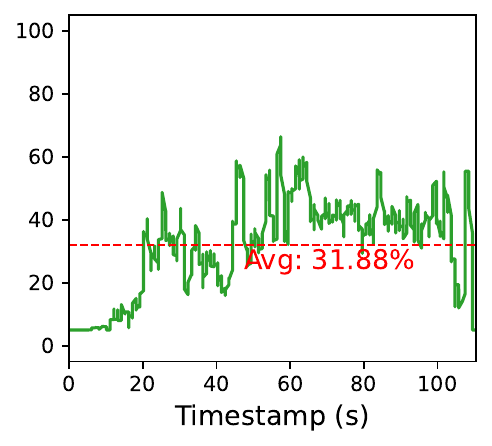}
        \caption{\naivemix}
        \label{fig:naivemix_8}
    \end{subfigure}
    \hfill
    \begin{subfigure}[t]{0.19\textwidth}
        \centering
        \includegraphics[height=90pt]{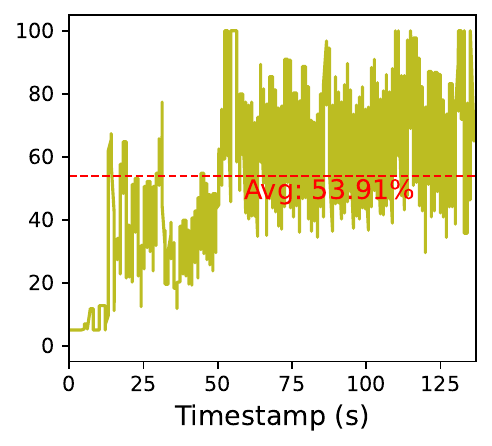}
        \caption{\approach}
        \label{fig:lemix_8}
    \end{subfigure}
    \caption{GPU utilization of three \separate setups, \naivemix, and \approach when deploying Llama-8B on eight A100 GPUs under LMSYS workloads. \separate (2-2) and \separate (1-3) dedicate 2 (1) nodes to inference and 2 (3) nodes to training, while \separate (dynamic) alternates between these configurations based on request rates. \naivemix and \approach co-locate both workloads across all four nodes.}
    \vspace{-10pt}
    \label{fig:mix_sep_profiling}
\end{figure*}

Figure~\ref{fig:sep_4_4} empirically demonstrates inefficiencies of the \separate strategy running a Llama-8B model~\cite{touvron2023llama} across four nodes (servers) with eight A100 GPUs.
Each model instance is divided into two stages on a node, with two nodes allocated to serving inference requests based on real-world traces from the LMSYS Chatbot Arena~\cite{zhenglmsys, zheng2023judging}, and the remaining two nodes dedicated to training.
We observe that inference suffers from low average GPU utilization at just 13.62\% due to inactive request periods.
Despite a higher utilization of 29.16\%, distributed training exhibits interleaving patterns caused by pipeline idle intervals. 

\noindent\textbf{Joint inference and training.} To address these inefficiencies, we first propose \naivemix, a strategy that \emph{co-locates} training and inference workloads on shared nodes via a simple round-robin (RR) policy.
As shown in Figure~\ref{fig:naivemix_8}, \naivemix increases the average GPU utilization by around 8\% over \separate through dynamically interweaving inference and training tasks.
This increased utilization arises from the overlap of inference and training workloads, enabling otherwise idle GPU resources to be utilized more efficiently.
Moreover, \naivemix enables \emph{on-the-fly} adaptation for concurrent workloads, where training updates immediately enhance the accuracy of subsequent co-located inference tasks, eliminating the delay associated with periodic inter-node model synchronization~\cite{fu2024serverlessllm} required in \separate setups.

However, na\"{\i}ve co-location alone offers limited gains, with \naivemix achieving an average of 31.88\% GPU utilization. 
One reason is that \naivemix is not sufficiently \emph{fine-grained}, leading to co-execution interference between asynchronous workloads. For example, unfinished training backward passes can race and delay subsequent inference requests, especially under high traffic.
Furthermore, real-world workloads exhibit \emph{dynamic} request arrival patterns~\cite{sun2024llumnix}, with \emph{heterogeneous} request lengths~\cite{zhong2024distserve,agrawal2024taming} that exacerbate execution delays and additional GPU idle periods.
During high traffic, these inefficiencies compound as memory contention induces prolonged overruns, causing degraded responsiveness and potential service level objective (SLO) violations~\cite{sheng2024fairness}.

\noindent\textbf{A comprehensive solution.}
Recognizing these challenges, we propose \approach, a fine-grained framework for co-operating LLM training and inference workloads in distributed systems. 
\approach aims to:
(1) maximize resource utilization, (2) improve inference accuracy via continuous retraining, and (3) meet response time SLO.
Unlike \naivemix, \approach integrates task-specific execution awareness and runtime adaptability to dynamically balance these objectives. 

Specifically, \approach leverages an 
offline profiler to gather latency and memory coefficients on each hardware (\S\ref{sec:offline_profiling}).
These profiled results enable the system to speculate the execution behaviors of online tasks and their system-wide impact (\S\ref{sec:execution_prediction}), while also providing actionable insights for co-optimizing multiple objectives during resource allocation (\S\ref{sec:task_allocation}) and runtime execution scheduling (\S\ref{sec:execution_scheduling}).
As a result, \approach can \emph{consolidate} workloads onto fewer nodes during periods of light demand—such as low request rates 
or reduced training intensity—without compromising SLO compliance.
As shown in Figure~\ref{fig:lemix_8}, \approach significantly improves the average GPU utilization by around 22\% over \naivemix, through fine-grained node partitioning and runtime scheduling based on real-time workload conditions (e.g., request rate, retraining burden).
This adaptability ensures \approach remains robust in real-world scenarios with dynamic arrival patterns and heterogeneous computational demands. 


\noindent\textbf{Contributions.} 
We evaluate \approach on diverse models and hardware, including three GPT models on A6000 GPUs and three Llama models on A100 GPUs, leveraging model parallelism under both synthetic and real workload traces.
\approach delivers up to 3.53$\times$ higher throughput, 0.61$\times$ lower inference loss, and 2.12$\times$ higher SLO attainment over \separate under various conditions.
Our contributions are:
\begin{itemize}[nosep,leftmargin=*]
    \item To our knowledge, this is the first study of concurrent LLM training and inference in distributed systems, unveiling inefficiencies in both \separate (\S\ref{sec:separate}) and \naivemix (\S\ref{sec:mix_motivation}) setups and identifying optimization opportunities. 
    \item We propose \approach, a dynamic task-specific framework that coordinates mixed workloads, optimizing resource utilization, inference accuracy, and response SLO compliance under diverse real-time conditions (\S\ref{sec:methodology}).
    \item We extend the scope of co-location techniques to support a wide range of LLM serving (e.g., autoregressive generation) and parallel training (e.g., data parallelism) scenarios, demonstrating the generality of \approach (\S\ref{sec:discussion}).
\end{itemize}

\section{Background and Analysis of \separate}
\label{sec:separate}

Recent surge in deploying LLMs for continuous user interaction~\cite{shen2024large} has necessitated effective strategies for handling concurrent training and inference tasks. 
These workloads span responding to user queries and retraining aimed at aligning LLM outputs with human preferences~\cite{christiano2017deep,li2025dr} and ensuring factuality~\cite{dai2023safe,fu2024safety}.
A widely adopted approach to manage such workloads in distributed systems is the \separate strategy, which dedicates servers in different phases to either inference or training tasks~\cite{choi2021multi}. While this setup simplifies management and ensures inference accuracy, it exhibits resource inefficiencies under realistic workload conditions.

\subsection{System Model and Terminology}

\begin{table}[h]
    \centering
    \resizebox{0.5\textwidth}{!}{
    \small
    \begin{tabularx}{0.5\textwidth}{>{\hsize=0.08\hsize}X>{\hsize=0.95\hsize}X}
    \toprule
        Stage & Sharded sub-model (e.g., layers 0-8) on a specific GPU. \\
        Node & A single server containing multiple GPUs (e.g., 8$\times$A100). \\
        Task & An inference or training operation across stages within a node, where an inference task is forward-only, while a training task includes a forward + backward. \\
        Batch & Internal mini-/micro-batching strategies within each task. They do not affect how \approach views or schedules tasks across nodes. \\
    \bottomrule
    \end{tabularx}
    }
    \caption{List of key terms used in the paper.}
    \label{tab:notations}
    \vspace{-0.3cm}
\end{table}

Table~\ref{tab:notations} includes a system model, defining key terms such as task, node, and stage of execution in this paper. We assume tasks are independent (no inter-task communication), and their arrivals are externally triggered (e.g., by user queries).

\subsection{Empirical Analysis of \separate}
Figure~\ref{fig:sep_2_6} highlights the limitation of \separate under another static ``1-3'' partition, allocating one node for inference and three nodes for training.
While this setup slightly increases the average inference utilization to 14.28\%, training utilization drops to 27.78\% compared to the ``2-2'' partition, leading to negligible overall gains.
These observations suggest that altering the fixed partition of nodes alone provides limited benefits.

To address workload variability, we further evaluate a more adaptive \emph{dynamic} \separate strategy that alternates between ``1-3'' and ``2-2'' setups based on request rates (``1-3'' for less than 50 rps and ``2-2'' otherwise), as shown in Figure~\ref{fig:sep_dynamic}.
However, even with these adjustments, the average utilization only improves marginally to 26.43\%, indicating the persistent inefficiencies of partitioning regardless of node allocation.
These inefficiencies stem from two key challenges: serving idleness due to dynamic request arrivals and training pipeline idleness caused by workload heterogeneity.

\subsection{Two Main Sources of Inefficiencies in \separate}
\noindent\textbf{Serving idleness in dynamic environments.}
Modern LLM deployments operate in a \emph{multitenant} environment with highly \emph{dynamic} request patterns across multiple node instances~\cite{sun2024llumnix}. 
To address the significant memory demands of LLM inference, model parallelism~(MP)~\cite{jia2019beyond} and continuous batching~\cite{yu2022orca} are widely employed, wherein each GPU hosts a sharded model slice (\refer stage) and sequentially executes forward passes of a mini-batch to reduce latency through statistical multiplexing~\cite{li2023alpaserve}. 
While MP enables serving large models on memory-constrained devices, it exacerbates GPU idle periods during light traffic. 
For example, under low request rates in Figure~\ref{fig:separate_illustration}, idle GPU intervals, such as between tasks 2 and 3 on Node 1, are frequent, leaving significant portions of the pipeline underutilized.
This inefficiency becomes more pronounced in large-scale distributed systems, where per-node request rate is even lower during off-peak hours~\cite{gptai2024servers}.

\begin{figure}[]
    \centering
    \includegraphics[width=0.48\textwidth]{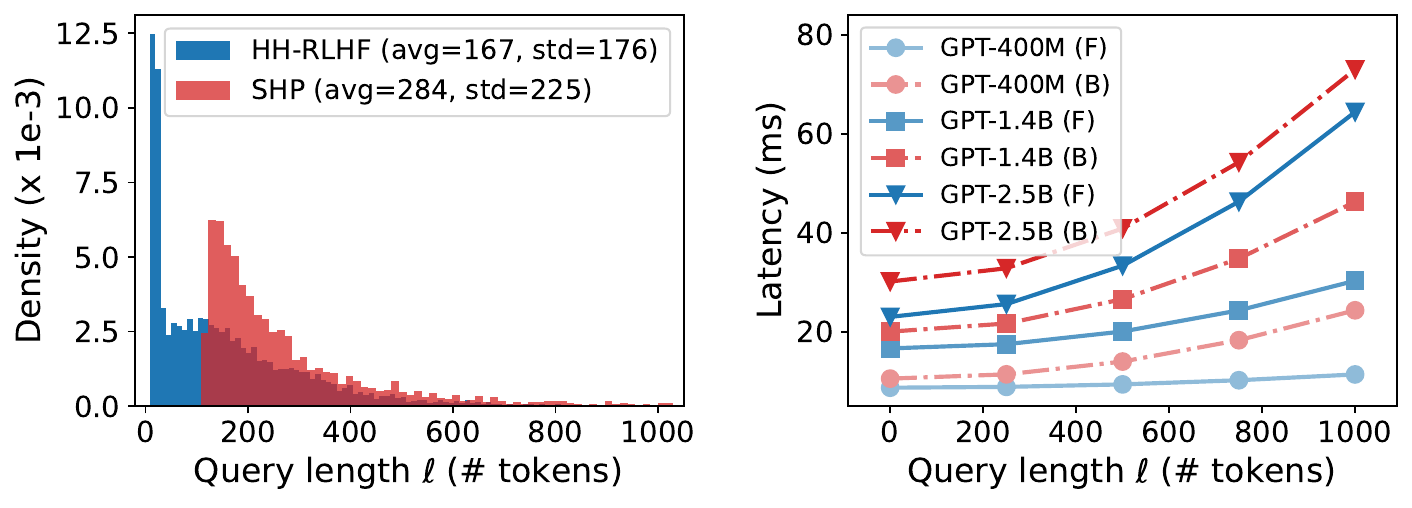}
    \caption{\emph{Left:} The length distribution (w/ standard deviation) of two datasets and \emph{Right:} the forward (F) and backward (B) latency running GPT models on a single RTX A6000 GPU.}
    \label{fig:latency_study}
    \vspace{-10pt}
\end{figure}

\noindent\textbf{Pipeline idleness from workload heterogeneity.}
To meet the high computational costs of LLM training, pipeline parallelism~(PP)~\cite{huang2019gpipe,athlur2022varuna} extends MP to divide mini-batches into smaller micro-batches that are processed in parallel to maximize utilization. 
While asynchronous PP (A-PP)~\cite{gaunt2017ampnet} theoretically minimizes idleness by continuously processing micro-batches without flushing, LLM tasks with context-specific inputs introduce a unique challenge: \emph{workload heterogeneity}.
As shown in Figure~\ref{fig:latency_study}, real-world user preference alignment datasets: HH-RLHF~\cite{bai2022training} and SHP~\cite{ethayarajh22a}, exhibit significant variability in query lengths, leading to diverse forward and backward latencies due to attention computation~\cite{vaswani2017attention}.
This variability disrupts the seamless execution of A-PP, as overlapping forward and backward passes contend for GPU resources, leading to execution delays and increased idle periods.
For example, on Node 2 of Figure~\ref{fig:separate_illustration}, task 2$'$ (\raisebox{0.2ex}{\textcolor{black}{\scalebox{1.0}{$\bigtriangledown$}}}) begins execution on GPU2 (S2) much later due to interference from the backward pass of task 1$'$.
Similarly, task 5$'$ (\raisebox{0ex}{\textcolor{black}{\scalebox{1.0}{$\bdiamond$}}}) cannot leverage the pipeline idle periods left by task 3$'$ or task 4$'$.
This postponed inter-stage execution is termed as \emph{far dependency}~\cite{choi2023envpipe}, as opposed to \emph{immediate dependency} when tasks have the same lengths and execute without delays.
These inefficiencies accumulate over time, resulting in prolonged E2E latency and fragmented memory. 
Although increasing concurrency, such as splitting mini-batches into more micro-batches, partially mitigates idleness, the associated communication overhead~\cite{athlur2022varuna} and far dependencies impose fundamental limits on system efficiency.


\section{Motivation of Workload Co-Location}
\label{sec:mix_motivation}

\begin{figure*}[htbp]
    \centering
    \begin{subfigure}[t]{0.95\textwidth}
        \centering
        \includegraphics[width=\textwidth]{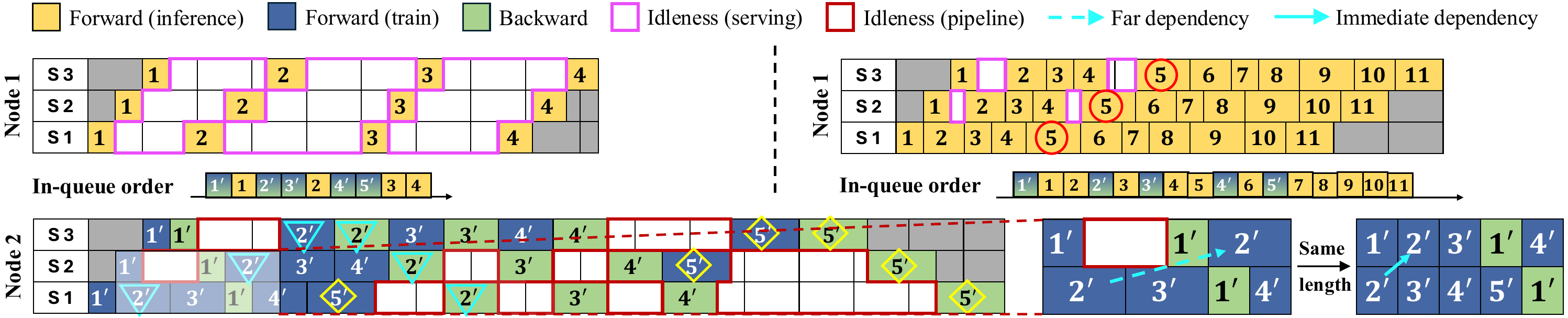}
        \caption{\separate operates inference workloads under low (left) and high (right) request rates on Node 1 and training workloads on Node 2.}
        \label{fig:separate_illustration}
    \end{subfigure}
    \\
    \vspace{10pt} 
    \begin{subfigure}[t]{0.95\textwidth}
        \centering
        \includegraphics[width=\textwidth]{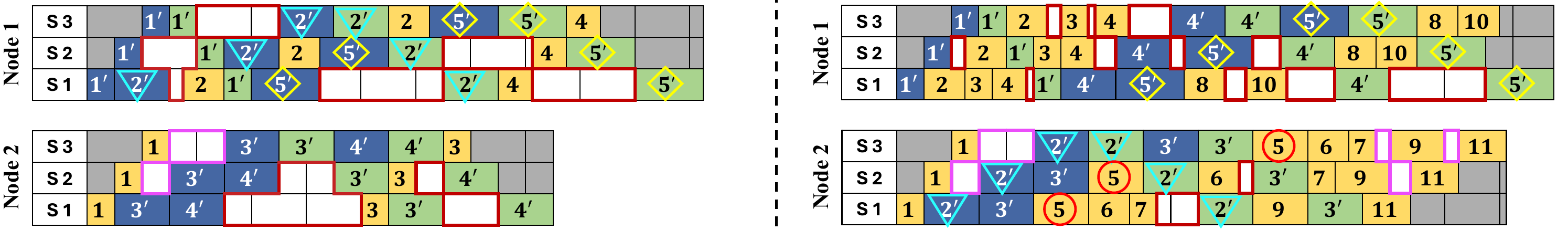}
        \caption{\naivemix co-locates both workloads by assigning tasks in a Round-Robin~(RR) manner based on their in-queue order.}
        \label{fig:mix_illustration}
    \end{subfigure}
    \caption{Comparison of (a) \separate and (b) \naivemix under \emph{Left}: low request rates and \emph{Right}: high request rates on a two-node cluster.
    For simplicity, we present one micro-batch for each mini-batch.}
    \label{fig:mix_opportunity}
    \vspace{-10pt}
\end{figure*}

The evolving in-context learning capabilities of LLMs~\cite{wei2022chain} present an opportunity to optimize resource utilization by learning while serving inference requests. 
We propose that \emph{co-locating} training and inference workloads on shared resources mitigate the inefficiencies inherent in \separate.
A baseline strategy, \naivemix, assigns tasks to nodes using a fair Round-Robin (RR) policy based on their in-queue order, as shown in Figure~\ref{fig:mix_illustration}.
Inference tasks are queued according to their online arrival times, while training tasks are queued according to the first-stage (S1) forward end times of previous training tasks. 
For example, in the right part of Figure~\ref{fig:mix_opportunity}, the first four tasks enqueued are 1$'$, 1, 2, 2$'$. Consequently, \naivemix co-locates tasks 1$'$ and 2 on Node 1, while allocating tasks 1 and 2$'$ to Node 2 for execution. 

\begin{figure}[t]
    \centering
    \begin{subfigure}[t]{0.49\linewidth}
        \centering
        \includegraphics[height=90pt]{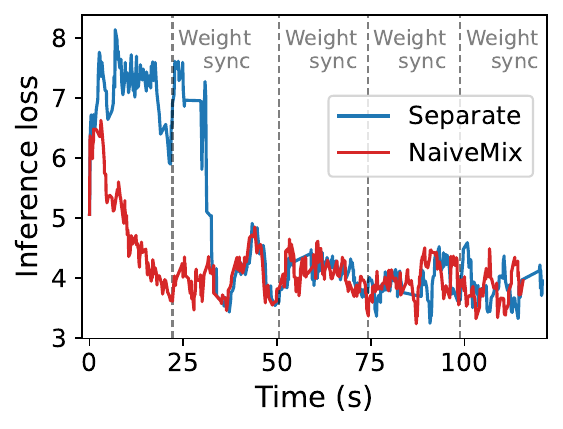}
        \caption{Continuous vs. periodic retrain}
        \label{fig:loss_overtime}
    \end{subfigure}
    \hfill
    \begin{subfigure}[t]{0.49\linewidth}
        \centering
        \includegraphics[height=90pt]{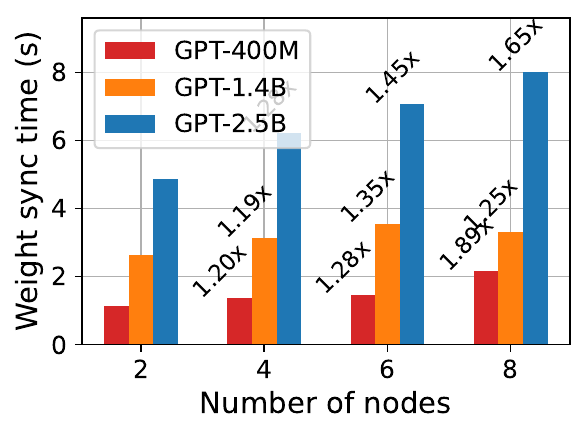}
        \caption{Inter-node sync latency}
        \label{fig:sync_overhead}
    \end{subfigure}
    \caption{Real data showcasing (a) inference loss over time and (b) weight synchronization latency across nodes for \separate.}
    \label{fig:separate_overhead}
    \vspace{-10pt}
\end{figure}

\subsection{Advantages of \naivemix}
\label{sec:naivemix_benefits}
\noindent\textbf{De-fragmentation.} 
\naivemix effectively enhances resource utilization compared to \separate by reducing idle periods across various real-time conditions:
\begin{itemize}[nosep,leftmargin=*]
    \item At low request rates, \naivemix alleviates serving idleness by running training workloads within the arrival gaps. For example, majority of training tasks 3$'$ and 4$'$ on Node 2 are executed in the interval between inference tasks 1 and 3. 
    \item At high request rates, \naivemix reduces pipeline idleness by inserting inference workloads into underutilized stages of training pipelines. For example, task 8 on Node 1 utilizes the idle pipeline period left by tasks 4$'$ and 5$'$.
\end{itemize}

\noindent\textbf{Latency reduction.}
By distributing training tasks across potentially more nodes, \naivemix naturally relaxes the far dependencies caused by forward-backward interference in A-PP. For example, task 5$'$ completes much earlier than in \separate training.
These optimizations in utilization consequently reduces E2E latency—the cumulative execution time across all nodes for executing a given workload, enabling more GPU time to be devoted to actual computations.

\noindent\textbf{Enhanced serving quality.}
\naivemix continuously improves inference accuracy by co-locating both workloads, which allows model instances to be updated in near real-time, benefiting serving quality on shared nodes.
For example, under low traffic, the performance of task 4 on Node 1 is enhanced due to updates after the backward pass of training task 1$'$.
Conversely, as shown in Figure~\ref{fig:loss_overtime}, \separate requires periodically synchronizing (e.g., checkpointing) updated weights~\cite{fu2024serverlessllm} from training to inference nodes, sacrificing serving quality before each communication. 
The synchronization overhead is also non-trivial and grows with model size and cluster scale (e.g., the number of nodes), as illustrated in Figure~\ref{fig:sync_overhead}.

\subsection{Optimization Opportunities}
\label{sec:naivemix_optimization}

While \naivemix alleviates multi-source inefficiencies of \separate, its effectiveness is limited due to coarse-grained task-agnostic execution awareness.

\begin{figure}[t]
    \centering
    \begin{subfigure}[t]{0.49\linewidth}
        \centering
        \includegraphics[height=88pt]{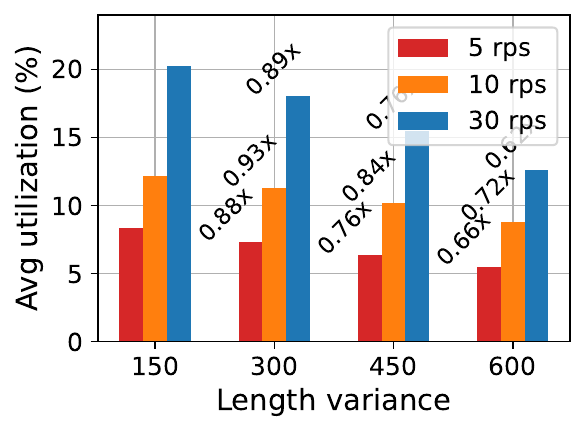}
        \caption{Workload heterogeneity}
        \label{fig:LH-vs-idle}
    \end{subfigure}
    \hfill
    \begin{subfigure}[t]{0.49\linewidth}
        \centering
        \includegraphics[height=88pt]{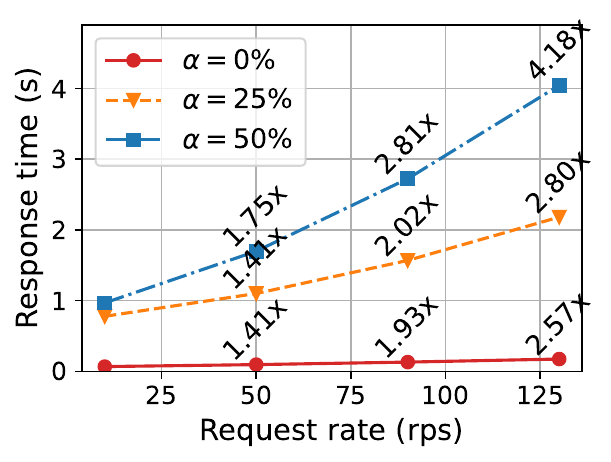}
        \caption{Traffic intensity}
        \label{fig:lambda-vs-response}
    \end{subfigure}
    \caption{Impact of (a) workload heterogeneity on utilization under different request rates and (b) request rate on serving responsiveness under different training rates $\alpha$.}
    \vspace{-10pt}
    \label{fig:length_results}
\end{figure}

\noindent\textbf{Suboptimal resource utilization.}
\naivemix still leaves spaces for utilization optimization.
For example, low \emph{training rate}—the proportion of training tasks in the queue—limits the reduction in serving idleness when insufficient training tasks are available to reduce the serving idleness if distributed across more nodes, such as between tasks 2 and 4 on Node 1 in Figure~\ref{fig:mix_illustration} under low request rates.
Moreover, workload heterogeneity remains a challenge.
In Figure~\ref{fig:mix_illustration}, task 1 is assigned to Node 2 under low request rates.
However, allocating task 1 to Node 1 would instead utilize the idle periods left by task 1$'$. 
Figure~\ref{fig:LH-vs-idle} validates this impact on \naivemix's utilization across subsets with different length variances sampled from HH-RLHF~\cite{bai2022training}. 
We observe a consistently decreasing GPU utilization under highly-variable query lengths, particularly under heavy traffic.   
These inefficiencies highlight the need for a \emph{fine-grained} policy that accounts for task-specific contributions to GPU idle periods during resource allocation.

\noindent\textbf{Prolonged serving response time.}
Na\"{\i}ve co-location inevitably enlarges inference response times due to resource contention between training and serving, especially during high-traffic periods. 
In Figure~\ref{fig:mix_illustration}, task 5 (\raisebox{0.12ex}{\textcolor{black}{\scalebox{1.0}{$\bigcirc$}}}) on Node 2 takes longer to complete compared to its execution on Node 1 under \separate. 
Our empirical results of deploying GPT-2.5B under \naivemix in Figure~\ref{fig:lambda-vs-response} show a prolonged inference response time as request and training rates grow, due to more frequent co-execution interference. 
A fine-grained approach is needed to estimate task-specific response times and \emph{prioritize} inference tasks that risk violating SLO targets.

\noindent\textbf{Memory contention in overloaded systems.}
\naivemix lacks runtime awareness of resource availability, leading to potential memory contention~\cite{wu2023transparent} when concurrent workloads interfere in their pipelines.
Under heavy traffic, autoregressive inference workloads may overlap with computation-intensive backward passes, especially when longer sequences and KV cache~\cite{pope2023efficiently} are involved. 
These simultaneous workloads can trigger memory overruns and further extend latencies, which is compounded by the unpredictable behaviors of accelerators under high load, such as non-deterministic scheduling delays and memory fragmentation~\cite{reagen2017methods}.
Therefore, a runtime profiler is needed to monitor resource usage, defer or offload memory-blocking workloads to ensure SLO compliance.

\section{Design of \approach}
\label{sec:methodology}

\begin{figure}[]
    \centering
    \includegraphics[width=0.47\textwidth]{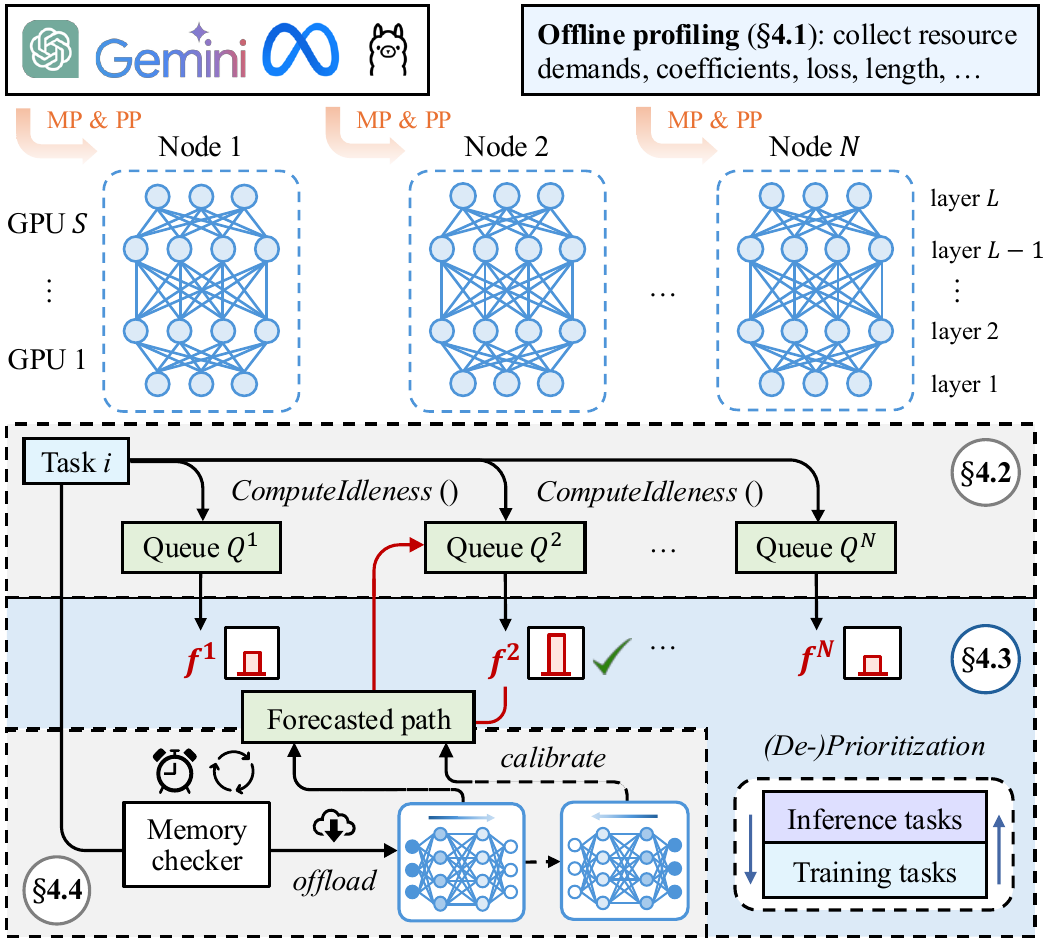}
    \caption{\approach's components and their interactions.}
    \label{fig:overview}
    \vspace{-10pt}
\end{figure}

From \S\ref{sec:mix_motivation}, we discuss that LLM systems with concurrent workloads benefit from joint training-inference and can be further optimized by task-specific scheduling.
We propose \approach, a fine-grained framework for co-operating LLM training and inference workloads in distributed systems, addressing inefficiencies such as serving idleness and pipeline delays. 
As shown in Figure~\ref{fig:overview}, \approach first conducts \textbf{offline profiling} to derive latency- and memory-sensitive coefficients for estimating hardware-dependent execution latencies (\S\ref{sec:offline_profiling}). 
Using these profiled results, \approach speculates the idle impact and serving latency of incoming tasks on each node through \textbf{task-specific execution planning} (\S\ref{sec:execution_prediction}).
These predictions aid in determining task preferences for each node, which are integrated with global task prioritization for \textbf{hierarchical resource (node) allocation} to balance utilization and SLO compliance (\S\ref{sec:task_allocation}).
After assigning tasks, \approach executes them while mitigating memory contention through \textbf{runtime scheduling}, selectively offloading caches, and balancing the memory demands of concurrent workloads (\S\ref{sec:execution_scheduling}).

\subsection{Offline Profiling} 
\label{sec:offline_profiling}
To support efficient scheduling and memory-aware execution, \approach starts with an \emph{offline profiler} that captures the latency and resource characteristics of LLM workloads. 

\noindent\textbf{Latency estimation.}
Forward and backward execution latencies scale quadratically with \emph{query length} $\ell$ (Figure~\ref{fig:latency_study}) and linearly with \emph{batch size} $C$. 
We model Stage-level execution time as $\Delta_F = \eta_{F}^n\cdot C \cdot \ell^2, ~\Delta_B = \eta_{B}^n \cdot C\cdot \ell^2$, where $\eta_{F}^n$ and $\eta_{B}^n$ are hardware-dependent coefficients. 
These are obtained by measuring average stage runtimes across various batch sizes\footnote{For serving requests, we apply iteration-level FCFS continuous batching~\cite{yu2022orca} to form inference mini-batches.} and query lengths from the HH-RLHF~\cite{bai2022training} and SHP~\cite{ethayarajh22a} datasets, normalized by $C \cdot \ell^2$.

\noindent\textbf{Memory profiling.}
To preempt memory saturation, we profile peak GPU memory usage under synthetic workloads with increasing $\ell$ and $C$.
To balance throughput and stability, \approach defines a memory utilization threshold, $M_{\text{threshold}} = \kappa \cdot M_{\text{peak}}$, where $\kappa$ is a safety factor. 
We select $\kappa$ empirically to avoid memory overflows while preserving throughput and responsiveness under diverse real-time conditions.
When memory constraints prevent immediate execution, tasks are delayed in the queue. However, excessive delay impacts responsiveness. Offline profiler empirically varies the maximum wait time $T_{\text{max}}$ and identify a cutoff beyond which task latency degrades sharply. 
This value bounds deferrals before offloading.

\noindent\textbf{Insights for online scheduling.}
The results of offline profiling are integrated into \approach for online scheduling and resource management. Latency coefficients ($\eta_F^n$, $\eta_B^n$) enable fine-grained planning of forward and backward executions (\S\ref{sec:execution_prediction}), while memory thresholds ($M_{\text{threshold}}$) and waiting tolerance limits ($T_{\text{max}}$) guide adaptive queueing and runtime scheduling decisions (\S\ref{sec:execution_scheduling}).
Additionally, offline profiler offers practical insights in \emph{continuous retraining strategies}. By correlating training rates with model accuracy and SLO requirements, practitioners can adaptively adjust system priorities:
\begin{itemize}[nosep, leftmargin=*]
    \item If performance metrics (e.g., loss) are acceptable offline or SLO deadline is stringent, the system reduces the training rate during online to prioritize inference throughput. 
    \item Conversely, resources can be prioritized for training to pursue accuracy gains while attaining SLO (\S\ref{sec:task_allocation}).
\end{itemize}

\subsection{Task-Specific Execution Planning} 
\label{sec:execution_prediction}

\begin{figure}[]
    \centering
    \includegraphics[width=0.49\textwidth]{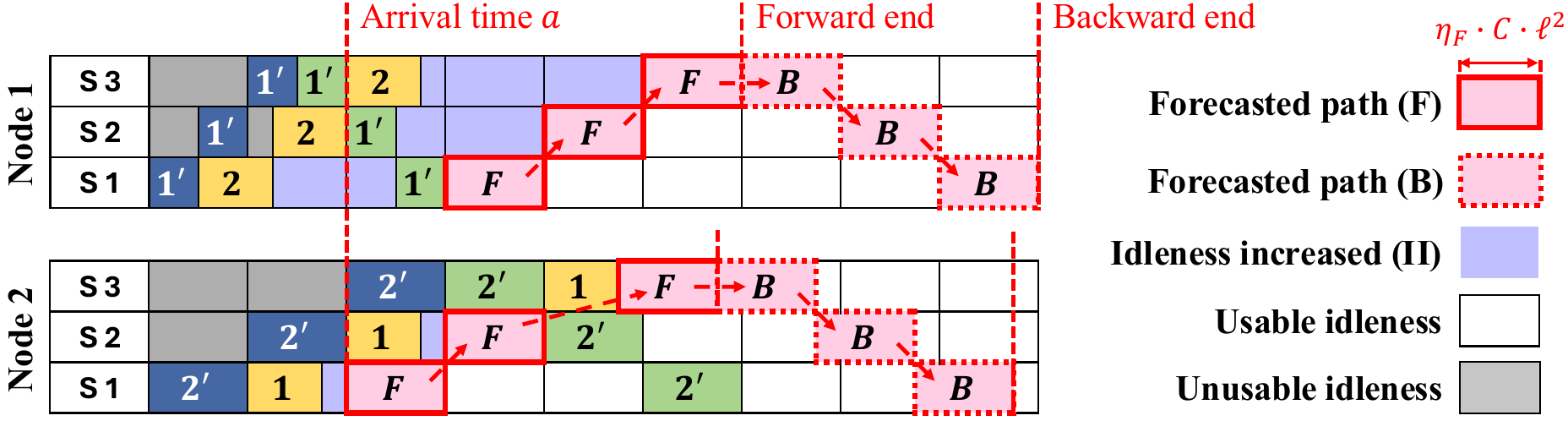}
    \caption{An illustration of how \approach estimates task-specific II (\crule[blue!50!white!50!]{4mm}{3mm}{-0.3ex}) and response time R based on arrival time $a$, length $\ell$, and batch size $C$ through execution planning. Dashed lines indicate that backward operations are omitted for inference tasks.}
    \label{fig:task_assignment}
    \vspace{-10pt}
\end{figure}

When a task arrives in a distributed LLM system, the first challenge is to assess its impact if executed on a specific node.
This involves anticipating both resource utilization and response latency introduced by the task.
To address \naivemix's limitations (\S\ref{sec:naivemix_optimization}), \approach adopts fine-grained execution planning of the task by considering its arrival (enqueued) time, execution latencies, and inter-stage dependencies, as shown in Figure~\ref{fig:task_assignment}. Specifically:
\begin{itemize}[nosep,leftmargin=*]
    \item \emph{Idleness increased} (II): \approach distinguishes between ``usable'' and ``unusable'' idle periods on each node. Task-specific II measures the resource utilization gap created by the new task's execution, which transforms previously ``usable'' idle periods into ``unusable'' ones.
    \item \emph{Response time} (R): the estimated latency from the task's arrival time to the completion of its forward pass.
\end{itemize}
Their calculation is detailed in Algorithm~\ref{alg: predict_response_idle}.

\noindent\textbf{Initialization.}
For each node $n$, \approach maintains two trace queues $Q_{\text{train}}^n$ and $Q_{\text{inference}}^n$, which track the execution paths of ongoing tasks.
The forward path of a new task is initialized based on the end time of the preceding task $\emph{task}_{\text{prev}}$ and the immediate dependency.
For each stage $s$, its forward start time is the later of two events: its forward end time in the preceding stage $s-1$ or $\emph{task}_{\text{prev}}$'s forward end time in the current stage (line 5). The forward end time is incremented by an estimated forward execution latency (line 6).

\begin{algorithm}[t]
\small
\caption{\textsc{ComputeIdleness}}
\label{alg: predict_response_idle}
\begin{algorithmic}[1]
\State \textbf{Input:} number of GPUs per node $S$,~trace queues $Q^n := Q_{\text{train}}^n \cup Q_{\text{inference}}^n$ for each node $n$, new \emph{task} batch: $\{a, \ell, C\}$
\State Parameters: forward and backward coefficient $\eta_{F}^n, \eta_{B}^n$ 
\State Initialize: $\text{II}\gets 0, ~\emph{task}_{\text{prev}} \gets Q^n[-1]$, $Q_{\text{temp}} \gets Q_{\text{train}}^n$
\For {\textnormal{each stage} $s \in \{1 ... S\}$}
  \State $\emph{task}.\text{start}_{f}^{s} \gets \max(\emph{task}.\text{end}_{f}^{s-1}, \emph{task}_{\text{prev}}.\text{end}_{f}^{s})$
  \State $\emph{task}.\text{end}_{f}^{s} \gets \emph{task}.\text{start}_{f}^{s} + \eta_{F}^n\cdot C\cdot\ell^2$
  \State $\emph{offset} \gets 0$ 
  \While {$Q_{\text{temp}} \neq \emptyset$}
    \State $\emph{task}_{\text{train}}: \{a_{\text{train}}, \ell_{\text{train}}, C_{\text{train}}\} \gets Q_{\text{temp}}$.dequeue()
    \If{ $\emph{task}.\text{end}_{f}^{s} \leq \emph{task}_{\text{train}}.\text{start}_{b}^{s}$}
      \State $Q_{\text{temp}} \gets Q_{\text{temp}} \cup \{\emph{task}_\text{train}\}$ 
      \State \textbf{break} 
    \EndIf
    \State $\emph{task}.\text{start}_{f}^{s} \gets \max(\emph{task}.\text{start}_{f}^{s}, \emph{task}_{\text{train}}.\text{end}_{b}^{s})$
    \State $\emph{task}.\text{end}_{f}^{s} \gets \emph{task}.\text{start}_{f}^{s} + \eta_{F}^n\cdot C\cdot\ell^2$ 
    \If{ 
    $\emph{task}_{\text{prev}}.\text{end}_{f}^{s} \leq \emph{task}_{\text{train}}.\text{start}_{b}^{s}$ 
    }
      \State $\emph{offset} \gets \emph{offset} + \eta_{B}^n\cdot C_{\text{train}}\cdot\ell_{\text{train}}^2 $ 
    \EndIf
    \If{ 
    $s = 1 \wedge \textsc{CheckExecuted}(\emph{task}_{\text{train}}, s)$ 
    }
      \State Remove $\emph{task}_{\text{train}}$ from $Q_{\text{train}}^n$
    \EndIf 
  \EndWhile
  \State $\text{II} \gets \text{II} + \emph{task}.\text{start}_{f}^{s} - \emph{task}_{\text{prev}}.\text{end}_{f}^{s} - \emph{offset}$ 
\EndFor
\State $\text{R} \gets \emph{task}.\text{end}_{f}^{S} - a$
\State \textbf{return} II, R
\end{algorithmic}
\end{algorithm}

\noindent\textbf{Far dependency rescheduling.} 
To efficiently account for far dependencies caused by co-execution interference, \approach introduces a temporary queue, $Q_{\text{temp}}$, which tracks pending training tasks in $Q_{\text{train}}^n$.
For each training task in $Q_{\text{temp}}$:
\begin{itemize}[nosep,leftmargin=*]
    \item If the current forward can finish before the backward of the training task begins, the stage executes the forward pass during this idle pipeline period. The task is then reinstated in $Q_{\text{temp}}$ for subsequent stages (lines 10-12).
    \item Otherwise, the forward start time is postponed to the backward end time of this interfered task, and the forward end time is recalculated accordingly (lines 13-14). Particularly, if this training task completes its backward pass, it is removed from the node queue $Q_{\text{train}}^n$ to avoid recomputation in subsequent stages (lines 17-18). 
\end{itemize}
All backwards executed between the previous and current forward (e.g., task 1$'$ on Node 1 in Figure~\ref{fig:task_assignment}) are recorded, and their cumulative execution times are subtracted from the serving intervals to yield II (line 15-19). The response time R is estimated after forward path planning (line 20).

\noindent\textbf{Backward planning.}
For tasks requiring training, the backward paths are initialized immediately after the forward paths. The backward start and end times are sequentially computed for each stage in reverse order, propagating back through the pipeline and maintaining a dependency chain between stages until the traversal ends. The planned backward paths aid in forecasting subsequent tasks before the execution.

\subsection{Hierarchical Resource (Node) Allocation}
\label{sec:task_allocation}

Based on anticipated idle periods and response times for each node, \approach optimizes resource allocation to address three objectives—maximizing resource utilization, minimizing serving response time, and improving serving quality under dynamic workload conditions. 
To achieve this, \approach employs a two-level hierarchical policy:
\begin{itemize}[nosep,leftmargin=*]
    \item \emph{Task-level} (local): New tasks are assigned to nodes with the highest priority scores, calculated using heuristic balancing the three objectives. For example, the new task in Figure~\ref{fig:task_assignment}, if assigned to Node 2, would produce lower II and R. 
    \item \emph{Queue-level} (global): \approach dynamically adjusts task priorities in the global task queue, deprioritizing training tasks that risk delaying subsequent inference tasks (and violating SLO goals) at high request rates.  
\end{itemize}

\begin{figure}[]
    \centering
    \includegraphics[width=0.485\textwidth]{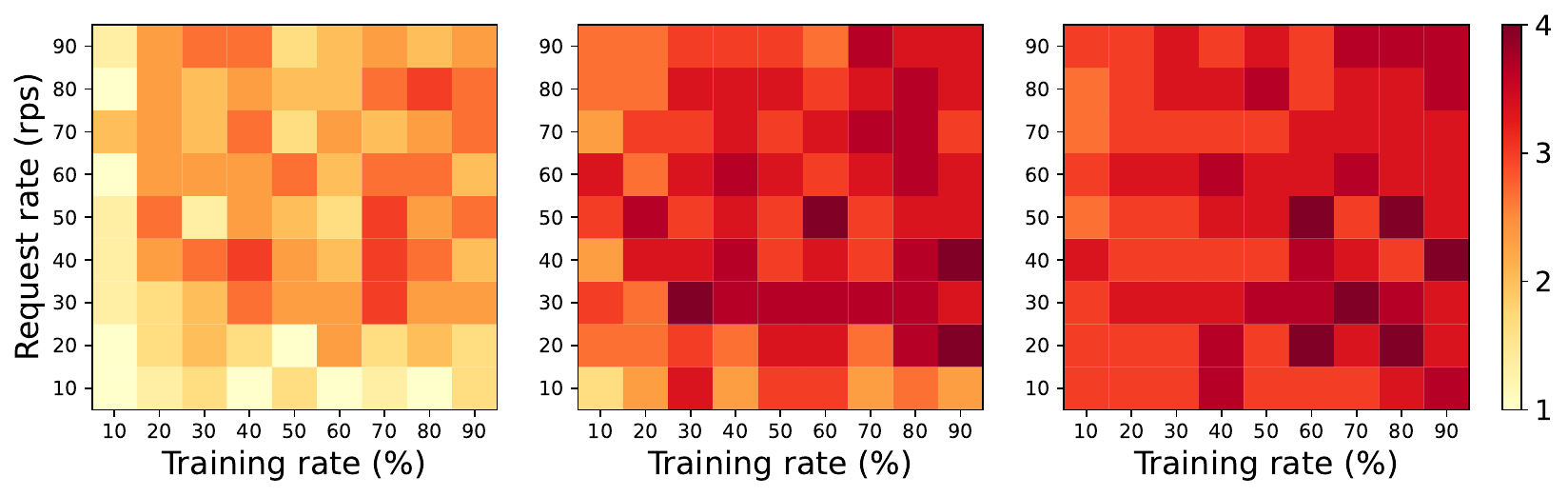}
    \begin{minipage}[t]{0.15\textwidth} 
        \centering
        \vspace{-10pt}
        \caption*{\footnotesize(a) GPT-400M}
        \label{fig:nodes_gpt_400m}
    \end{minipage}
    \hfill
    \hspace{-5pt}
    \begin{minipage}[t]{0.15\textwidth}
        \centering
        \vspace{-10pt}
        \caption*{\footnotesize(b) GPT-1.4B}
        \label{fig:nodes_gpt_1.4b}
    \end{minipage}
    \hfill 
    \hspace{-5pt}
    \begin{minipage}[t]{0.15\textwidth}
        \centering
        \vspace{-10pt}
        \caption*{\footnotesize(c) GPT-2.5B}
        \label{fig:nodes_gpt_2.5b}
    \end{minipage}
    \hspace{10pt}
    \vspace{-5pt}
    \caption{Number of allocated nodes (out of 4) for \approach when processing concurrent workloads under various setups.}
    \vspace{-10pt}
    \label{fig:active_nodes}
\end{figure}

\noindent\textbf{Task-level multi-objective node allocation.}
\approach defines \emph{idleness profit} (IP) to capture the utilization benefit of accommodating the incoming task to a specific node
\begin{equation}
    \label{eq:idle_profit}
    \text{IP} = - \max \left \{ \frac{\text{II}}{S} - (a-a_{[-1]}), \tau \right \},
\end{equation}
where $a-a_{[-1]}$ represents the \emph{inter-arrival interval} between the preceding and incoming task, $S$ is the number of GPUs per node, and $\tau$ is a threshold.
IP penalizes nodes with large increased idle periods relative to arrival intervals, enabling \approach to adapt to dynamic request rates—a key characteristic of modern LLM-serving systems~\cite{sun2024llumnix}.

\approach incorporates \emph{length consistency}~(LC), which quantifies how well a task with query length $\ell$ aligns with a node's historical workload, as the serving quality heuristic
\begin{equation}
\label{eq:length_heterogeneity}
    \text{LC} = \frac{1}{\sigma_{<a}\sqrt{2\pi}}\exp{\left \{-\frac{(\ell - \mu_{<a})^2}{2\sigma_{<a)}^2}\right \}},
\end{equation}
where $\mu_{<a}$ and $\sigma_{<a}$ are the mean and standard deviation for previously executed tasks on the node. 
Higher LC scores align tasks with a node's workload profile, which has been shown benefiting training convergence~\cite{li2023rt,li2025mixtraining} and minimizing GPU idle periods caused by workload heterogeneity (\S\ref{sec:naivemix_optimization}). For inference tasks assigned to different nodes, \approach minimizes cache miss rate by prioritizing \emph{prefix reuse}~\cite{zheng2024sglang} for length-bucketed queries, which keeps most requests on warm caches.

The node priority score integrates the three objectives
\begin{equation}
\label{eq:node_fitness}
    f = \frac{\text{IP} + \lambda_2\cdot \text{LC}}{\lambda_1\cdot \text{R}},
\end{equation}
where $\lambda_1$ and $\lambda_2$ balance the trade-off between response latency R and serving quality heuristic. 
Tasks are assigned to nodes with the highest priority scores, dynamically updating the node queues.
This priority-based mechanism enables \emph{workload consolidation}, dynamically adjusting the number of active nodes based on real-time conditions, as shown in Figure~\ref{fig:active_nodes}. Key observations include:
\begin{itemize}[nosep,leftmargin=*]
    \item Higher training rates lead to an increased $\text{II}$, as more unfinished backward passes create contention for resources, reducing IP and $f$ for allocated nodes. This shift tasks to new nodes to alleviate resource contention.
    \item  As request rates rise, arrival intervals shorten, amplifying $\text{II}$ due to the added interference caused by workload heterogeneity, again lowering $f$ and favoring new nodes.
    \item Larger model sizes exacerbate workload heterogeneity, as they magnify $\eta_{F}^n$ and $\eta_{B}^n$ (Figure~\ref{fig:latency_study}). This increase in $\text{II}$ reduces the profitability (IP) of assigning tasks to allocated nodes, instead favoring new nodes.
\end{itemize}

\begin{figure}[]
    \centering
    \includegraphics[width=0.48\textwidth]{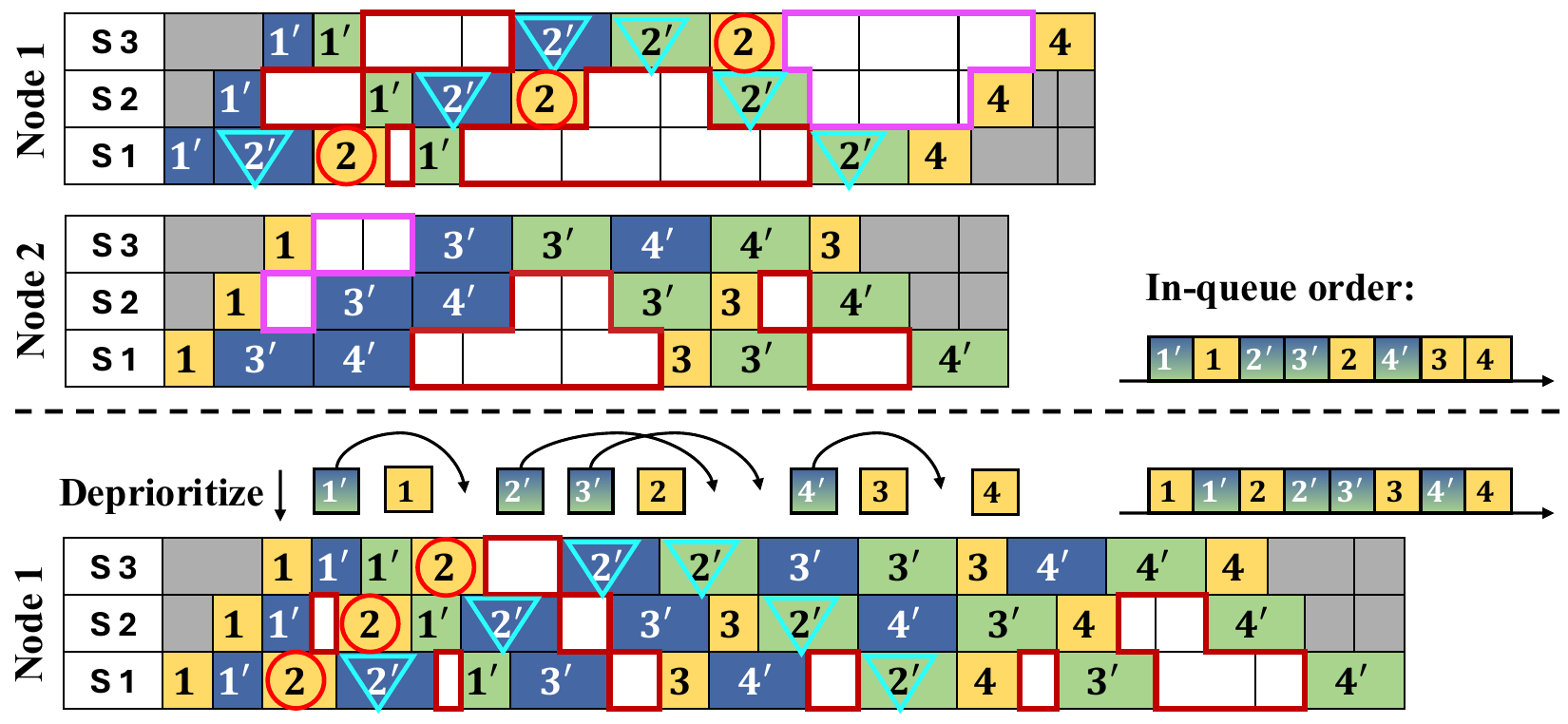}
    \caption{An illustration of how \emph{Bottom:} \approach consolidates workloads from \emph{Top:} \naivemix into one node to optimize utilization while maintaining SLOs via deprioritization.}
    \label{fig:lemix_opportunity}
    \vspace{-10pt}
\end{figure}

\noindent\textbf{Queue-level task prioritization.}
While local-level selection optimizes individual assignment, it neglects the inter-task impact that arises under high-traffic scenarios.
In such cases, both bursty inference requests and training tasks compete for shared resources, preempting subsequent inference tasks and risking SLO violations.
To maintain serving throughput, \approach \emph{deprioritizes} training tasks if the minimum estimated response time of its subsequent enqueued inference task $\emph{task}_{\text{next}}$ across all nodes violates the SLO target
\begin{equation}
    \min_{1\leq n\leq N }\{\max_{\emph{task}\in Q^n}\{\emph{task}_{\text{next}}.\text{end}_{f}^{S}\} + \eta_{F}^n\cdot {C'}\cdot{\ell'}^2\} - a' > \tau_{\text{R}},
\end{equation}
where $C'$, $\ell'$, $a'$ are the batch size, query length, and arrival time of $\emph{task}_{\text{next}}$. $\tau_{\text{R}}$ is a response SLO goal, e.g., 5$\times$ inference latency.
In Figure~\ref{fig:lemix_opportunity}, \approach defers training tasks 2$'$ and 3$'$ on Node 1, as either would significantly delay inference task 2.
These severe delays, if arise from far dependencies in preceding tasks (e.g., task 2$'$), also indicate under-utilization and could be optimized through this deprioritization process.
Practitioners can set different SLO goals to balance serving responsiveness and training efficiency. 
A larger $\tau_{\text{R}}$ increases delay tolerance, preserving training priority for improving serving quality, while a smaller $\tau_{\text{R}}$ penalizes SLO violations more strictly, prioritizing serving responsiveness.

\subsection{Runtime Memory-Aware Scheduling}
\label{sec:execution_scheduling}

After dispatching tasks to specific nodes, \approach coordinates all ongoing tasks with the new one on each node, ensuring reliable service under dynamic traffic. 
To handle the risk of memory overruns—especially critical when co-locating computation-intensive batched operations—\approach incorporates a runtime scheduler that maintains SLOs without sacrificing resource efficiency.
Memory overruns occur when overlapping pipelines exceed the system's available memory, potentially delaying inference generation or disrupting training operations. To preemptively manage memory and mitigate these issues, \approach employs a stage-level \emph{wait-or-drop} policy, as shown in Algorithm~\ref{alg:memory_aware_execution}. 

\begin{algorithm}[t]
\small
\caption{\textsc{ExecuteTaskMemoryAware}}
\label{alg:memory_aware_execution}
\begin{algorithmic}[1]
\State \textbf{Input:} 
number of GPUs per node $S$,~task queues $Q^n := Q_{\text{train}}^n \cup Q_{\text{inference}}^n$ for each node $n$. 
\State Parameters: Memory utilization threshold $M_{\text{threshold}}$, check interval $\Delta_t$, and maximum wait time $T_{\text{max}}$ 
\While {$Q^n \neq \emptyset$}
  \State $\emph{task} \gets Q^n$.dequeue() 
    \For {\textnormal{each stage} $s \in \{1 ... S\}$}
    \State $wait \gets 0$
    \While{\textbf{not} \textsc{MemoryAvailable}($n$, $s$, $M_{\text{threshold}}$)}
        \State $wait \gets wait + \Delta_t$
        \If{$wait \geq T_{\text{max}}$}
            \State  Offload KV cache of \emph{task} from GPU $s$
            \State \textbf{break}
        \EndIf
    \EndWhile
    \If{\textsc{MemoryAvailable}($n$, $s$, $M_{\text{threshold}}$)}
        \State \textsc{Forward}($\emph{task}$, $s$)
        \State Calibrate $<\emph{task}.\text{start}_{f}^{s}, \emph{task}.\text{end}_{f}^{s}> \in Q^n$ 
        \If{$s=S \wedge \emph{task}.\text{require\_backward}$}
            \State \textsc{Backward}($\emph{task}$)
            \State Calibrate $<\emph{task}.\text{start}_{b}^{*}, \emph{task}.\text{end}_{b}^{*}> \in Q_{\text{train}}^n$
        \EndIf 
    \EndIf
  \EndFor
\EndWhile
\end{algorithmic}
\end{algorithm}

\noindent\textbf{Wait-or-drop with execution calibration.}
Runtime monitoring is conducted before each forward on a stage. When an incoming task's memory demands exceeds the device's capacity due to concurrency, it is placed in a temporary wait state (lines 6-8). By deferring execution until enough memory is available, this phase prevents immediate conflicts without disrupting ongoing tasks. It is particularly beneficial for autoregressive generation tasks, where low latency is critical. If the memory remains constrained beyond a profiled threshold $T_{\text{max}}$, \approach \emph{offloads} intermediate activations and KV cache~\cite{pope2023efficiently} to CPU~\cite{eisenman2022check} to scale down memory demands and free resources for inference to attain SLOs (lines 9-10).
After execution, forecasted paths (\S\ref{sec:execution_prediction}) in the node queues are \emph{calibrated} by the execution traces (lines 13–17), facilitating future task planning and scheduling optimization. 

\section{Implementation and Discussion}
\label{sec:discussion}

\subsection{Implementation}
\approach is built on top of vLLM~\cite{kwon2023efficient} and DeepSpeed~\cite{aminabadi2022deepspeed} for MP and PP, and FlashAttention~\cite{dao2022flashattention} for memory optimization during inference generation. We realize asynchronous execution of stages using Python \emph{concurrent} library. 
Specifically, \approach employs a global scheduler that makes scheduling decisions oriented to the node instances, according to the priority and memory load of them. It invokes each node as a \emph{multi-threaded} execution environment, where each thread is assigned to manage a stage of the instance pipeline.
For each node, tasks are scheduled across these threads following the forward and backward dependencies. 
Each thread is responsible for coordinating and executing the forward and backward (if applicable) jobs for its assigned stage.
Once the memory availability for a task is confirmed, the appropriate thread invokes the execution process. 




\subsection{Model Update Synchronization}
In \separate, model on the final training node is periodically checkpointed (e.g., every 100 training tasks) and loaded onto inference nodes to ensure consistent serving quality.

Co-location methods, such as \naivemix and \approach, adopt a \emph{decentralized} strategy similar to federated learning where each node independently updates its local model instance based on its local training tasks. This setup resolves privacy concerns where sharing weight can be highly risky.

\subsection{Scalability and Distributed Architecture} 
We assume an implicit client-server architecture.
Real-world LLM systems (like Azure, AWS) do not rely on a centralized scheduler for every request. Instead, they adopt a distributed, hierarchical architecture that scales efficiently to millions of requests per minute.
LeMix operates on both \emph{cluster-level} and \emph{GPU-level}, with its global queue stored in a cluster, and each local queue stored in a server (node):
\begin{itemize}[nosep, leftmargin=*]
    \item \textbf{Front-end load balancers}: Distribute client requests geographically across server clusters (e.g., by regions, time zones) using stateless heuristics like round-robin or SLO-aware routing to handle high traffic volumes (e.g., 10k+ rps) without per-request scheduling.
    \item \textbf{Cluster-level scheduler}: \approach precisely allocate requests to specific servers within each cluster to coordinate training and inference tasks, which typically handles $<150$ rps. This aligns with our experimental setups and reflects the practical scale of resource allocation at the cluster level. As shown in analysis (\S\ref{sec:ablation_parameter_study}), \approach incurs negligible overhead under this workload.
    \item \textbf{GPU-level runtime scheduler}: Once tasks are assigned to a server, \approach manages execution using fine-grained techniques such as memory-aware batching, KV cache offloading, to optimize throughput and avoid OOM.
\end{itemize}


The efficiency of \approach is rooted in its lightweight design, enabling scalability in large-scale distributed systems. Task-specific execution planning involves forecasting idle periods and response times for incoming tasks. This step incurs an $O(S)$ complexity per node, where $S$ is the number of stages (GPUs) in the node. Resource allocation computes priority scores for all $N$ nodes and selects the optimal node for each task, resulting in an $O(N \cdot S)$ complexity. Memory-aware scheduling adjusts task execution based on runtime conditions such as memory availability and task inter-dependencies, maintaining an $O(S)$ overhead per node by only considering active queues.
Overall, \approach’s end-to-end scheduling operates with $O(N \cdot S)$ complexity across $N$ nodes and $S$ stages, ensuring efficient coordination even in large clusters.

\subsection{Autoregressive Generation}

\begin{algorithm}[t]
\small
\caption{\textsc{ContinuousBatching}}
\label{alg:continuous_batching}
\begin{algorithmic}[1]
\State \textbf{Input:} Maximum batch size $C$, maximum waiting time $T_w$, task queue $Q^n$ for a specific node $n$.
\State Initialize: batch $B^n \gets \emptyset$, batch start $T_{start} \gets \text{current time}$\;
\While{True}
    \If {$B^n = \emptyset$}
        \State $T_{start} \gets \text{current time}$
    \EndIf
    \While {$Q^n \neq \emptyset$ \textbf{and} $|B^n| < C$}
        \State $r \gets \text{get\_next\_request}(Q^n)$\;
        \If {not $r$.\text{require\_backward} $\wedge~ T_{start} + T_w > \text{current time}$}
            \State Add $r$ to $B^n$ with padding (if necessary)\;
        \Else
            \State \textbf{break}
        \EndIf
    \EndWhile
    \If {$B^n \neq \emptyset$}
        \State Execute batch $B^n$ on node $n$\;
        \State $B^n \gets \text{filter\_finished\_requests}(B^n)$\;
    \EndIf
\EndWhile
\end{algorithmic}
\end{algorithm}


Serving requests are handled using continuous batching into mini-batches up to a maximum size $C$ on a FCFS basis, as shown in Algorithm~\ref{alg:continuous_batching}. To prevent excessive delays during low-traffic periods, a maximum waiting time $T_w$ (defaulting to 0.5$\times$ inference latency) ensures timely execution even when the batch size $C$ is not reached. 
In particular, for autoregressive generation, we adopt hybrid iteration-level batching~\cite{yu2022orca}.
Each incoming request represents a prefilling workload, which is integrated with ongoing decoding workloads into mini-batches to improve throughput and responsiveness. 
In \approach, prefilling requests, which initialize new sequences, are allocatable across nodes based on our task assignment strategy, informed by execution predictions of continuously batched task on each node. 
Decoding requests, which extend prefilled contexts, operate continuously on the same node as their corresponding prefilling tasks, maintaining data locality and supporting dynamic scheduling, such as offloading, when memory bottlenecks arise.

\section{Evaluation}
\label{sec:experiments}

We evaluate \approach across different LLM sizes and workload conditions, including diverse request rates and training rates.
Experimental results show that \approach consistently outperforms baseline systems in all scenarios, achieving up to 3.53$\times$ higher throughput, 0.61$\times$ lower inference loss, and 2.12$\times$ higher SLO attainment compared to \separate (\S\ref{sec:synthetic_workloads}).
We further delve into node-level latency and response time breakdown under real workloads, showcasing how \approach's dynamic resource allocation enhances utilization and operational efficiency (\S\ref{sec:real_workload_experiments}).
The benefits of \approach become even more pronounced under high workload heterogeneity, where task-specific scheduling proves critical to performance gains (\S\ref{sec:lh_experiment}).
We also analyze fine-grained inference latency in different phases (\S\ref{sec:generation_latency}).
Finally, we conduct ablation studies to provide deeper insights into its techniques (\S\ref{sec:ablation_parameter_study}).

\begin{table}[]
    \centering
    \resizebox{0.485\textwidth}{!}{
    \begin{tabular}{lccccc}
    \toprule
        \textbf{Name} & \textbf{Size} & \textbf{\# Layers} & \textbf{Hidden size} & \textbf{Forward} & \textbf{Backward} \\
        \midrule
        GPT-400M & 1.0GB & 12 & 768 & 0.03s & 0.04s \\
        GPT-1.4B & 2.3GB & 24 & 1024 & 0.08s & 0.09s \\
        GPT-2.5B & 4.5GB & 36 & 1280 & 0.12s & 0.14s \\
        Llama-8B & 13GB & 32 & 4096 & 0.11s & 0.15s \\
        Llama-13B & 26GB & 40 & 5120 & 0.24s & 0.36s \\
        Llama-70B & 132GB & 80 & 8192 & 0.73s & 1.05s \\
        \bottomrule
    \end{tabular}}
    \caption{Model configuration and latency requirements. 
    }
    \label{tab:models}
    \vspace{-10pt}
\end{table}

\subsection{Experimental Setup}

\noindent\textbf{Datasets and models.}
We use two popular preference datasets targeting different alignment domains: HH-RLHF~\cite{bai2022training} (harmlessness) and SHP~\cite{ethayarajh22a} (helpfulness), to mimic the serving and alignment fine-tuning of LLM deployment.
Specifically, we sample 1,000 query-reference pairs in each online test to evaluate the model and system performance.
In this work, we consider GPT~\cite{zhang2020dialogpt} and Llama~\cite{touvron2023llama} family models of various sizes.
Table~\ref{tab:models} shows their detailed configurations.

\noindent\textbf{Testbed setups.}
We use two testbed configurations. 
For GPT models, we use 4 servers, each equipped with 2 NVIDIA RTX 6000 Ada GPUs (48 GB), 64 AMD EPYC-7543 CPU cores, and 2 TB memory. 
For larger Llama models, we use 4 AWS EC2 servers, each equipped with 2 NVIDIA A100 Tensor Core GPUs (80 GB), connected with pairwise NVLINK. 


\noindent\textbf{Synthetic workloads.} We mimic human requests as a Poisson process~\cite{li2023rt,li2023alpaserve,sheng2024fairness} and synthesize traces by sampling inter-arrival times from an exponential distribution with mean of the reciprocal of the various request rates. Tasks are sampled from the same dataset as training (e.g., SHP) to simulate online serving and retraining scenarios.

\noindent\textbf{Real workloads.} We further explore the effectiveness of \approach in more complex inference workloads constructed from the trace of LMSYS Chatbot Arena~\cite{zhenglmsys}—a real-world LLM serving platform for clients.

\noindent\textbf{Metrics.} 
We measure throughput as the average number of completed tasks per second and analyze the E2E latency across all nodes running the same amount of workloads, for utilization analysis.
For serving quality, we evaluate the average inference decoding loss.
For serving efficiency, we evaluate prefilling latency, i.e., time-to-first-token (TTFT), decoding latency, i.e., time-between-tokens (TBT)~\cite{zhong2024distserve},
and SLO attainment—proportion of inference tasks whose TTFT meet a SLO deadline of 5$\times$ forward latency~\cite{li2023alpaserve}.

\noindent\textbf{Baselines.}
We compare \approach with three baselines as below: 
\begin{itemize}[nosep,leftmargin=*]
    \item \separate~\cite{choi2023envpipe,li2023alpaserve}: Nodes are partitioned into training and inference nodes based on training rates $\alpha$. $N_{\text{train}} = \left \lfloor N\cdot\alpha+0.5 \right \rfloor$ nodes are dedicated to training and the remaining to inference to balance workloads. Each task is assigned to a training or inference node in a RR fashion.
    \item \textsc{Mix-RR} (\naivemix): Tasks are assigned to nodes in a sequential RR order. This method ensures a uniform distribution of tasks across all nodes.
    \item \utilmix: Lowest utilization first~\cite{feizabadi2003utilitiy}, serving as a strong baseline with more fine-grained execution awareness over RR, measures the GPU utilization and allocates tasks to nodes with the lowest average scores. \utilmix balances the active duration of all nodes over time, potentially preventing any single node from becoming a bottleneck. 
\end{itemize}



\begin{figure}[]
    \centering
    \includegraphics[width=0.5\textwidth]{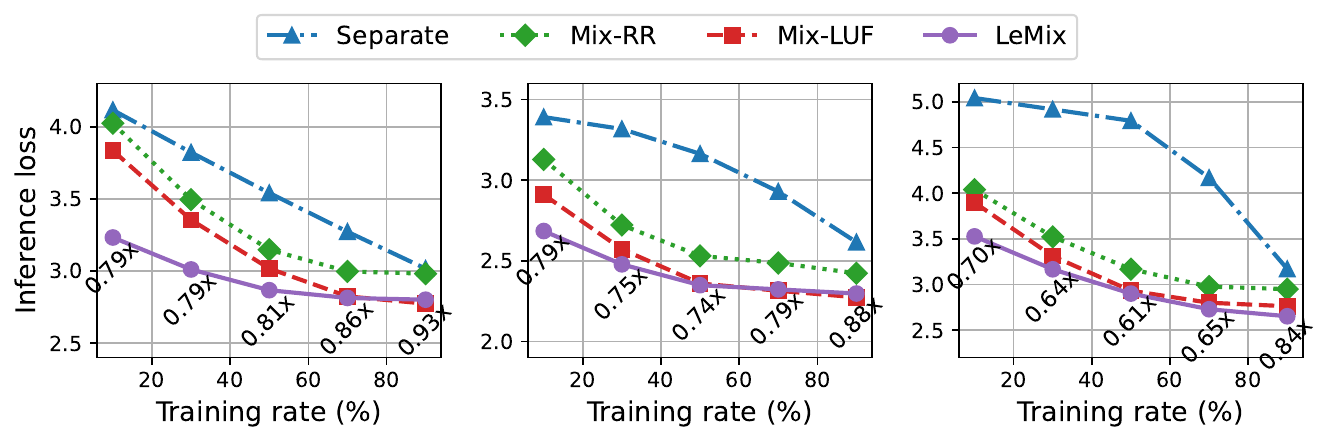}
    \begin{minipage}[t]{0.15\textwidth} 
        \centering
        \vspace{-10pt}
        \caption*{\footnotesize(a) Llama-8B}
    \end{minipage}
    \hfill
    \begin{minipage}[t]{0.15\textwidth}
        \centering
        \vspace{-10pt}
        \caption*{\footnotesize(b) Llama-13B}
    \end{minipage}
    \hfill 
    \begin{minipage}[t]{0.15\textwidth}
        \centering
        \vspace{-10pt}
        \caption*{\footnotesize(c) Llama-70B}
    \end{minipage}
    \vspace{-5pt}
    \caption{Average inference loss at various training rates.}
    \label{fig:loss_4nodes}
    \vspace{-10pt}
\end{figure}

\subsection{Results on Synthetic Workloads}
\label{sec:synthetic_workloads}

We assess \approach and the baselines in three dimensions—\emph{inference loss} of decoding human references to measure service quality under continuous retraining, \emph{throughput} as a resource utilization metric when running mixed workloads, and \emph{SLO attainment} to ensure prompt service responsiveness, under a spectrum of training rates for multiple models at different request rates on a four-node cluster.

\noindent\textbf{Improved serving quality under retraining.} 
Figure~\ref{fig:loss_4nodes} shows Llama models' average inference loss when running concurrent serving and training workloads.
We observe a general decreasing loss under larger training rates, as the model are retrained on more samples. All mixed approaches achieve a lower loss compared to \separate due to continuous updates from retraining (\S\ref{sec:naivemix_benefits}), with \approach consistently outperforming the others, reducing the average loss by up to 0.61$\times$ over \separate. This superiority indicates that our idea of workload co-location and consideration of LC (\S\ref{sec:task_allocation}) in \approach benefits the training procedure and convergence.
The discrepancy between \separate and co-location methods gets larger as model sizes grow, due to prolonged inter-node weight synchronization latencies (\S\ref{sec:naivemix_benefits}).

\begin{figure}[]
    \centering
    \includegraphics[width=0.5\textwidth]{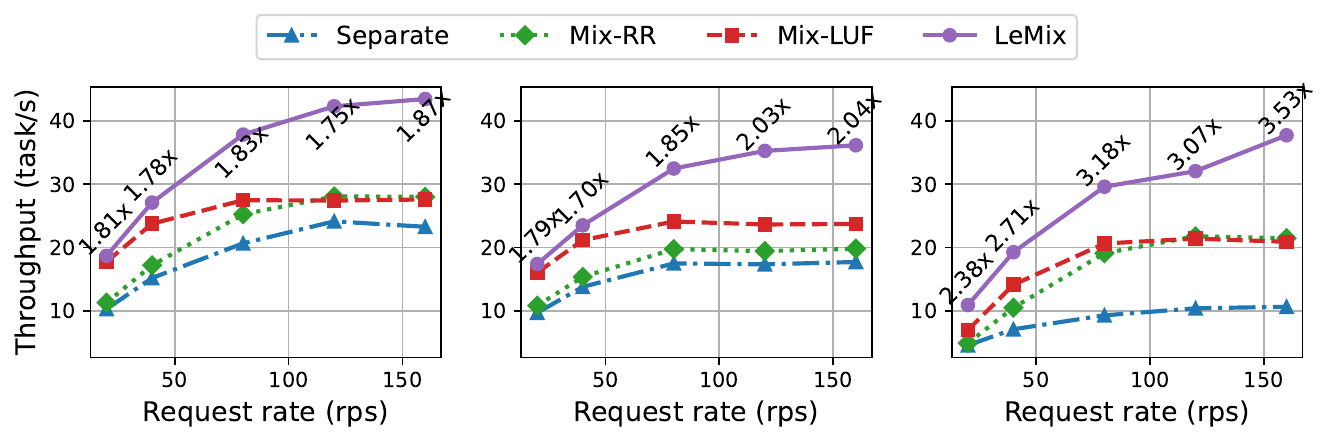}
    \begin{minipage}[t]{0.15\textwidth} 
        \centering
        \vspace{-10pt}
        \caption*{\footnotesize(a) Llama-8B}
    \end{minipage}
    \hfill
    \begin{minipage}[t]{0.15\textwidth}
        \centering
        \vspace{-10pt}
        \caption*{\footnotesize(b) Llama-13B}
    \end{minipage}
    \hfill 
    \begin{minipage}[t]{0.15\textwidth}
        \centering
        \vspace{-10pt}
        \caption*{\footnotesize(c) Llama-70B}
    \end{minipage}
    \vspace{-5pt}
    \caption{Throughput (task/s) across various request rates.}
    \vspace{-10pt}
    \label{fig:throughput_4nodes}
\end{figure}

\begin{figure}[]
    \centering
    \includegraphics[width=0.5\textwidth]{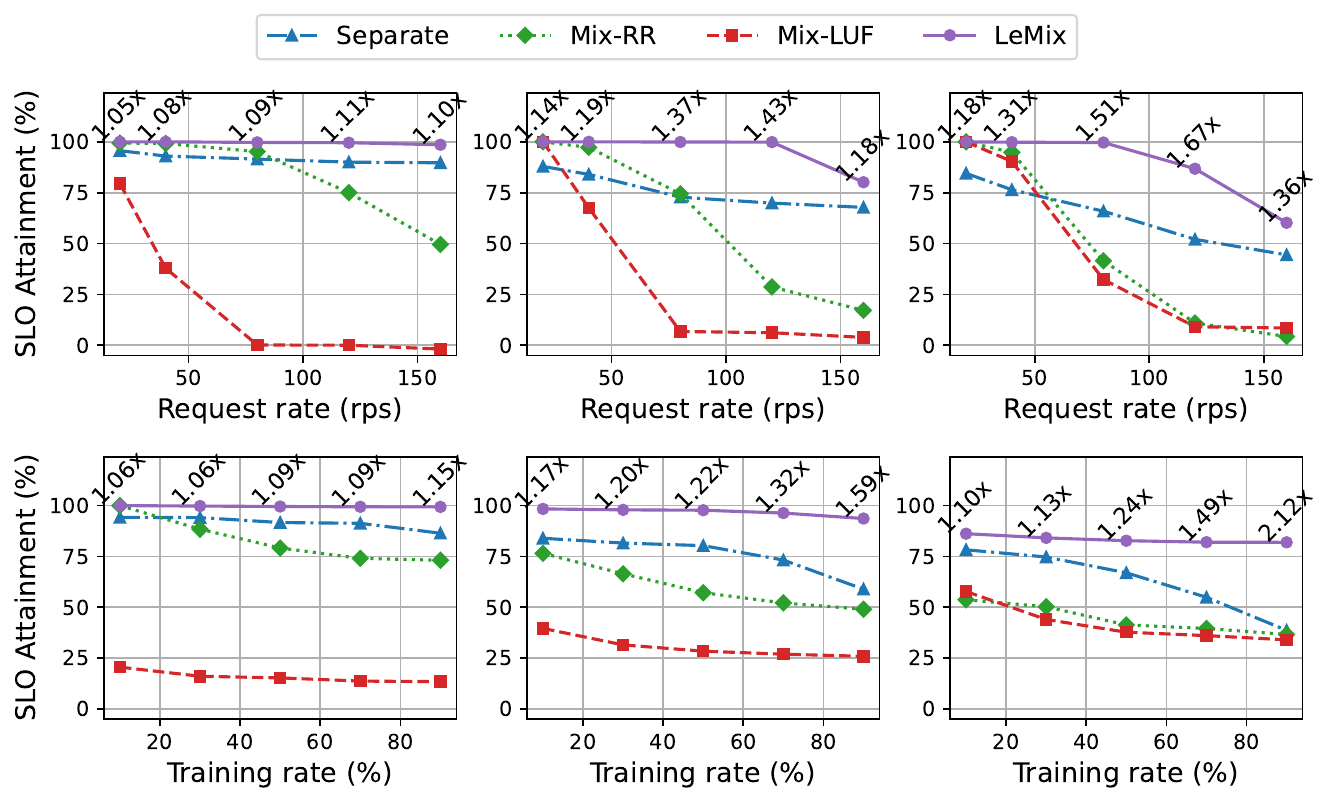}
    \begin{minipage}[t]{0.15\textwidth} 
        \centering
        \vspace{-10pt}
        \caption*{\footnotesize(a) Llama-8B}
        \label{fig:SLO_llama7b_4nodes}
    \end{minipage}
    \hfill
    \begin{minipage}[t]{0.15\textwidth}
        \centering
        \vspace{-10pt}
        \caption*{\footnotesize(b) Llama-13B}
        \label{fig:SLO_llama13b_4nodes}
    \end{minipage}
    \hfill 
    \begin{minipage}[t]{0.15\textwidth}
        \centering
        \vspace{-10pt}
        \caption*{\footnotesize(c) Llama-70B}
        \label{fig:SLO_llama70b_4nodes}
    \end{minipage}
    \vspace{-5pt}
    \caption{SLO attainment under various request and training rates.}
    \label{fig:SLO_4nodes}
    \vspace{-10pt}
\end{figure}

\noindent\textbf{Throughput under varying traffic intensity.} 
Figure~\ref{fig:throughput_4nodes} shows Llama models' throughput at various request rates when processing concurrent mixed workloads. We observe a notable enhance in resource efficiency for all methods at higher request rates, where \approach achieves the overall highest scores, improving throughput by up to 3.53$\times$ over \separate. 
Particularly, \utilmix also excels under light traffic conditions (e.g., less than 50 rps), which demonstrates the benefit of execution awareness in optimizing resource efficiency (\S\ref{sec:naivemix_optimization}), though such improvement over \naivemix diminishes in rapid task arrival due to the time-intensive GPU utilization querying procedure (Table~\ref{tab:profiling}) that significantly delays task execution. Moreover, the advantage of mixed methods over \separate is pronounced for larger models at high request rates, as the execution delays caused by workload heterogeneity are amplified under these circumstances (\S\ref{sec:separate}).

\noindent\textbf{SLO attainment in varying conditions.}
Figure~\ref{fig:SLO_4nodes} demonstrates a decreasing SLO attainment for Llama models as request rates (top row) and training rates (bottom row) increase. 
\approach consistently outperforms alternative methods, maintaining high SLO compliance across workload spectrum.
This resilience stems form its dynamic latency-aware scheduling which reduces response times for inference tasks.
In contrast, other mixed approaches suffer significant drop in SLO attainment, particularly at higher request rates, as co-locating workloads exacerbates contention for shared resources, delaying inference execution (\S\ref{sec:naivemix_optimization}).
While \separate achieves comparable performance to \approach under low workloads, its static property leads to declines in SLO compliance under higher training rates or request intensities.
The performance gap widens with larger model sizes (e.g., Llama-70B), where \approach achieves up to 2.12$\times$ better SLO attainment over \separate at high training rates, indicating its scalability in managing heterogeneous workloads. 

\subsection{Breakdown Analysis on Real Workloads}
\label{sec:real_workload_experiments}
We constructed real workload traces from LMSYS platforms following a similar process in \cite{sheng2024fairness}, where we treat each LLM as a client and sample requests from the trace and re-scale the real-time stamps to a time window.
During this serving window, training samples from the datasets are mixed to form in total 1,000 samples with varying training rates.
We evaluate node-level breakdown of \emph{E2E latencies} and \emph{response time} for \approach and the baselines.

\begin{figure}
    \centering
    \includegraphics[width=0.49\textwidth]{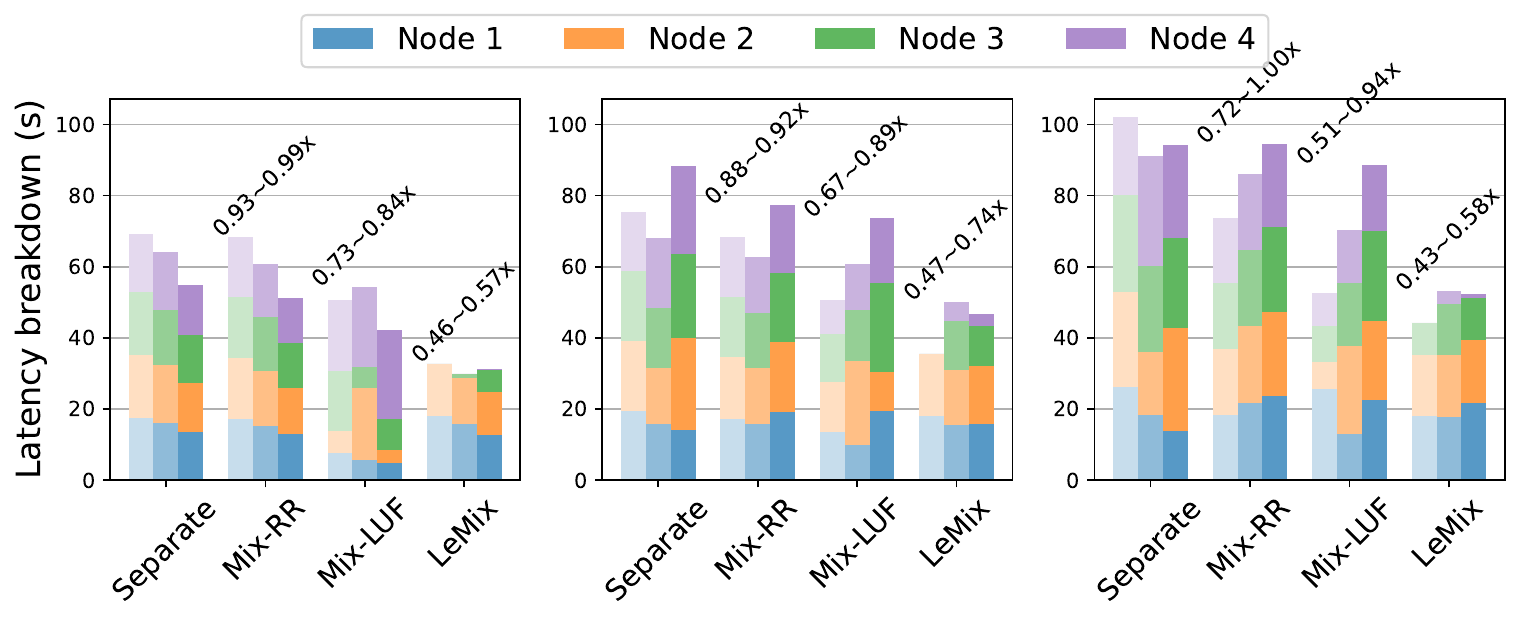}
    \begin{minipage}[t]{0.15\textwidth} 
        \centering
        \vspace{-10pt}
        \caption*{\footnotesize(a) GPT-400M}
    \end{minipage}
    \hfill
    \begin{minipage}[t]{0.15\textwidth}
        \centering
        \vspace{-10pt}
        \caption*{\footnotesize(b) GPT-1.4B}
    \end{minipage}
    \hfill 
    \begin{minipage}[t]{0.15\textwidth}
        \centering
        \vspace{-10pt}
        \caption*{\footnotesize(c) GPT-2.5B}
    \end{minipage}
    \vspace{-5pt}
    \caption{Breakdown E2E latencies under training rates of 10\% (light), 50\% (medium), and 90\% (dark color).}
    \label{fig:latency_node}
    \vspace{-10pt}
\end{figure}

\noindent\textbf{E2E latencies across nodes.} 
Figure~\ref{fig:latency_node} shows GPT models' E2E latencies breakdown on each node when processing real workloads. We observe a clear increased latency when running larger models due to more computational demands. All mixed approaches require less time than \separate to operate concurrent workloads, especially for larger models (e.g., GPT-2.5B) where workload heterogeneity amplifies the pipeline idleness of \separate (\S\ref{sec:separate}).
\approach achieves the lowest accumulative latencies, reducing the scores by up to 0.43$\times$ over \separate.
One notable reason behind such gain is its ability to ``dynamically'' adjust resource allocation (\S\ref{sec:task_allocation}) based on real-time conditions. For examples, \approach allocates only 2 nodes when running two smaller models under lower training rates (e.g., 10\%), effectively mitigating the sparsity issue in low workload demands.

\begin{figure}[]
    \centering
    \includegraphics[width=0.49\textwidth]{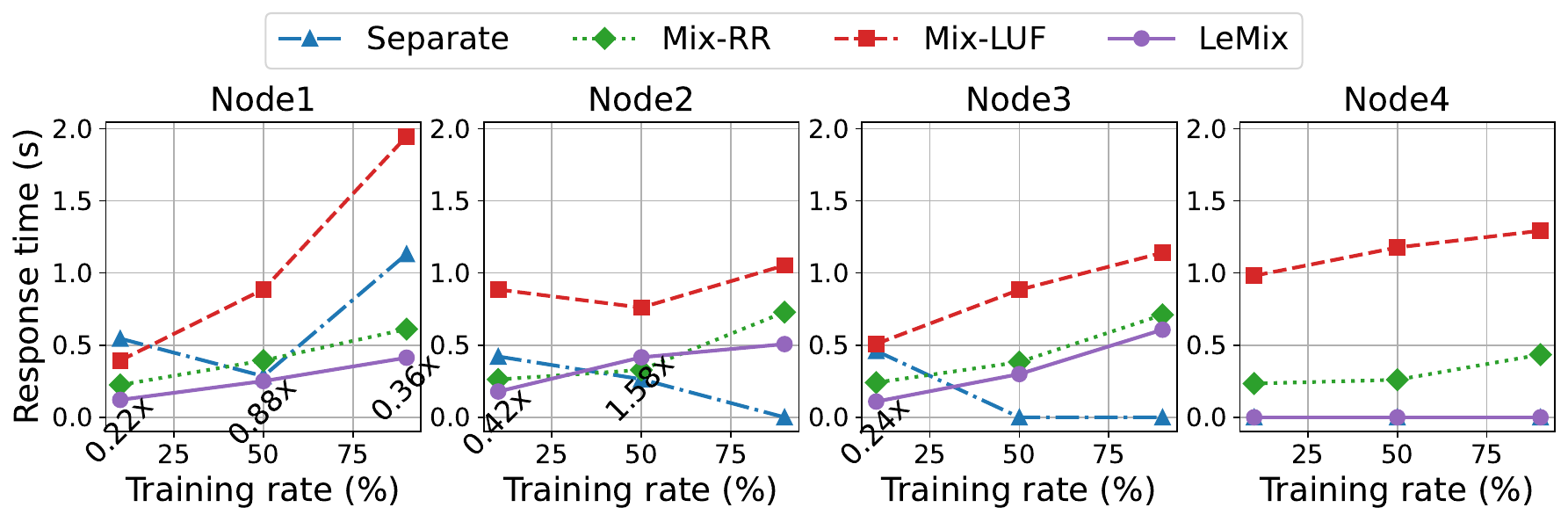}
    \begin{minipage}[t]{0.15\textwidth} 
        \centering
    \end{minipage}
    \hfill
    \begin{minipage}[t]{0.15\textwidth}
        \centering
    \end{minipage}
    \vspace{-10pt}
    \caption{Breakdown average response time on each node. Zero values mean no inference workloads are allocated on that node.}
    \label{fig:response_node}
    \vspace{-5pt}
\end{figure}

\noindent\textbf{Response time across nodes.} 
Figure~\ref{fig:response_node} details Llama-8B's average response time (TTFT) on four nodes across various training rates.
By default, Node 1 is dedicated to inference workloads in \separate, where \approach achieves lower response times, reducing the average scores by up to 0.22$\times$.
Such efficiency gain is more obvious under larger training rates where inference execution delays on Node 1 in \separate are amplified due to overloaded requests.
For the remaining three nodes, \approach still achieves the lowest response latencies under lower training rates (e.g., less than 50\%), while \separate excels under larger training rates since few or even no inference workloads are allocated to those nodes (e.g., Node 3).
\utilmix exhibits significant response delays due to the latency of querying GPU utilization, indicating its limited practical applicability when SLO attainment is pursued. 



\subsection{Impact of Workload Heterogeneity}
\label{sec:lh_experiment}


\begin{figure}
    \centering
    \includegraphics[width=0.48\textwidth]{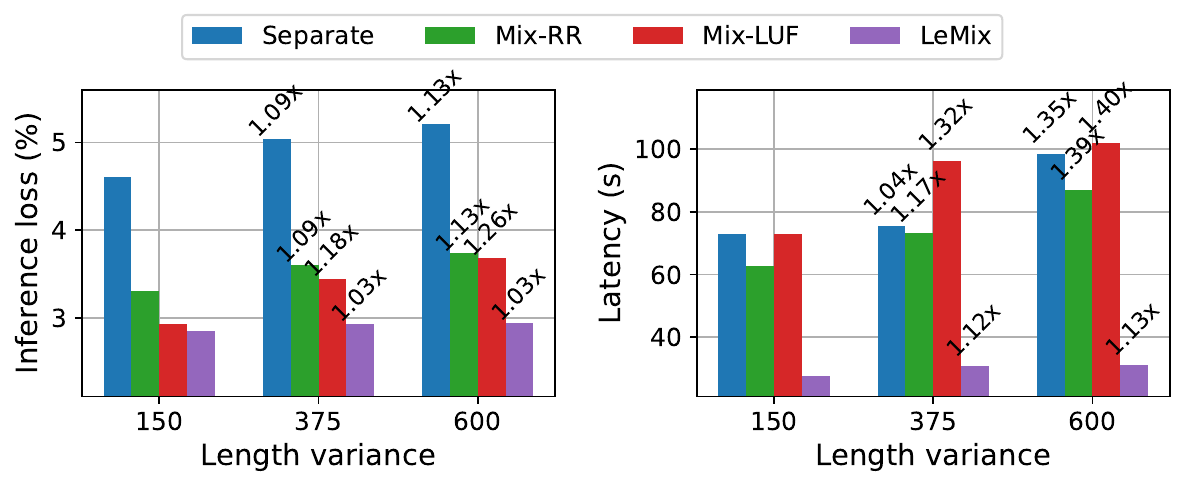}
    \caption{Impact of length heterogeneity on \emph{Left:} inference loss and \emph{Right:} E2E latencies under heavy traffic (150 rps).}
    \label{fig:length_heterogeneity_results}
    \vspace{-10pt}
\end{figure}

Rapid task arrival combined with larger workload heterogeneity could result in substantial idleness (\S\ref{sec:separate}) and hampered training convergence (\S\ref{sec:task_allocation}) when operating synthetic workloads. 
To explore how \approach handles this property inherent in real-world workloads, we simulate a range of length heterogeneities by sampling subsets from the two datasets.
We run GPT-2.5B model on each subset under a request rate of 150 rps across various training rates.
Figure~\ref{fig:length_heterogeneity_results} demonstrates a general increase in loss and E2E latency for \approach and the baselines as length heterogeneity grows. 
\approach is more robust against workload heterogeneity compared to the baseline methods: with up to 1.03$\times$ increase in inference loss and the maximum latency gain is 1.13$\times$.
This suggests that task-specific resource allocation can effectively mitigate the idleness incurred by co-execution interference and training inconsistency caused by workload heterogeneity.

\begin{figure}
    \centering
    \includegraphics[width=0.48\textwidth]{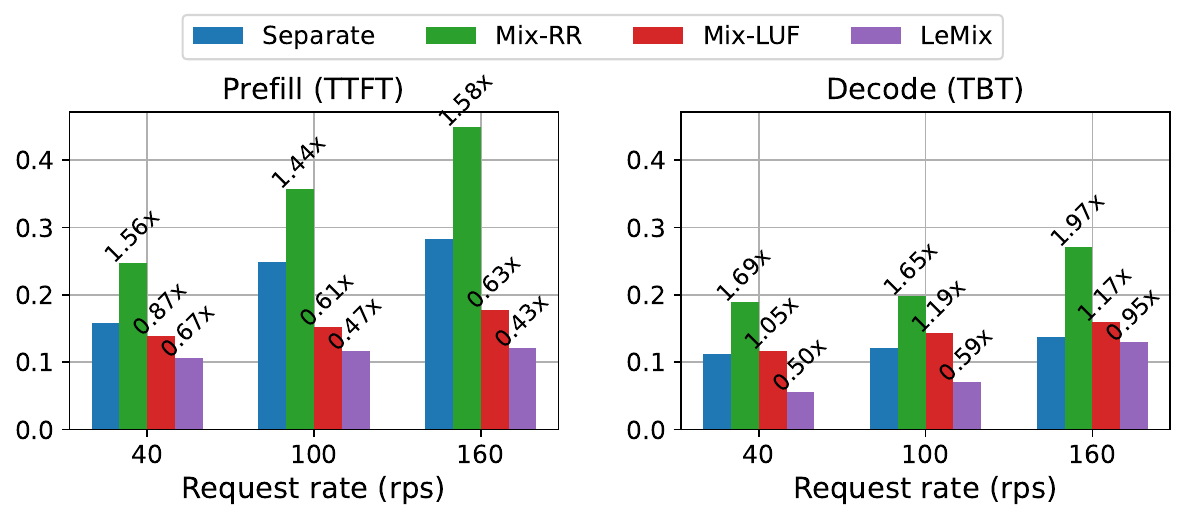}
    \caption{\emph{Left:} prefill and \emph{Right:} decode latency of inference serving under various request rates for GPT-2.5B.}
    \label{fig:latency(generate)_lambda}
    \vspace{-5pt}
\end{figure}

\subsection{Inference (Generation) Efficiency Study}
\label{sec:generation_latency}

To independently study inference latency under mixed training workloads, we run GPT-2.5B with a memory pool size of 1000 tokens allocated for KV cache~\cite{pope2023efficiently} across various request rates with 50\% training rate. 
Figure~\ref{fig:latency(generate)_lambda} highlights the performance variations across methods in both prefilling and decoding phases.
\approach consistently achieves the lowest latency in both TTFT and TBT across all request rates, achieving up to 0.43$\times$ lower TTFT and 0.50$\times$ lower TBT compared to \separate.
Both \utilmix and \naivemix incurs additional serving latency when co-locating serving and training workloads due to resource contention, and their TBT discrepancies are caused by the hybrid batches of both prefilling and decoding requests where node allocation (\S\ref{sec:task_allocation}) makes difference.
\utilmix exhibits relatively better TTFT performance as the computation-intensive prefilling workloads are allocated to low-utilization nodes, but its TBT latency becomes a bottleneck due to frequent GPU utilization tracking during decoding.
\approach dynamically allocates prefilling workloads across nodes (\S\ref{sec:task_allocation}) to prevent resource bottlenecks and cut TTFT latency, and leverages runtime scheduling (\S\ref{sec:execution_scheduling}) to fill idle GPU slots and use batching opportunities to minimize TBT during decoding. This adaptability ensures balanced resource utilization across computation- (prefill) and memory-bound (decode) phases, allowing \approach to excel in diverse conditions.

\subsection{Understanding \approach's Improvements}
\label{sec:ablation_parameter_study}

We run Llama-70B on synthetic workloads across varying request and training rates to evaluate the contributions of individual components in \approach and their overheads.

\begin{figure}[t]
    \centering
    \includegraphics[width=0.47\textwidth]{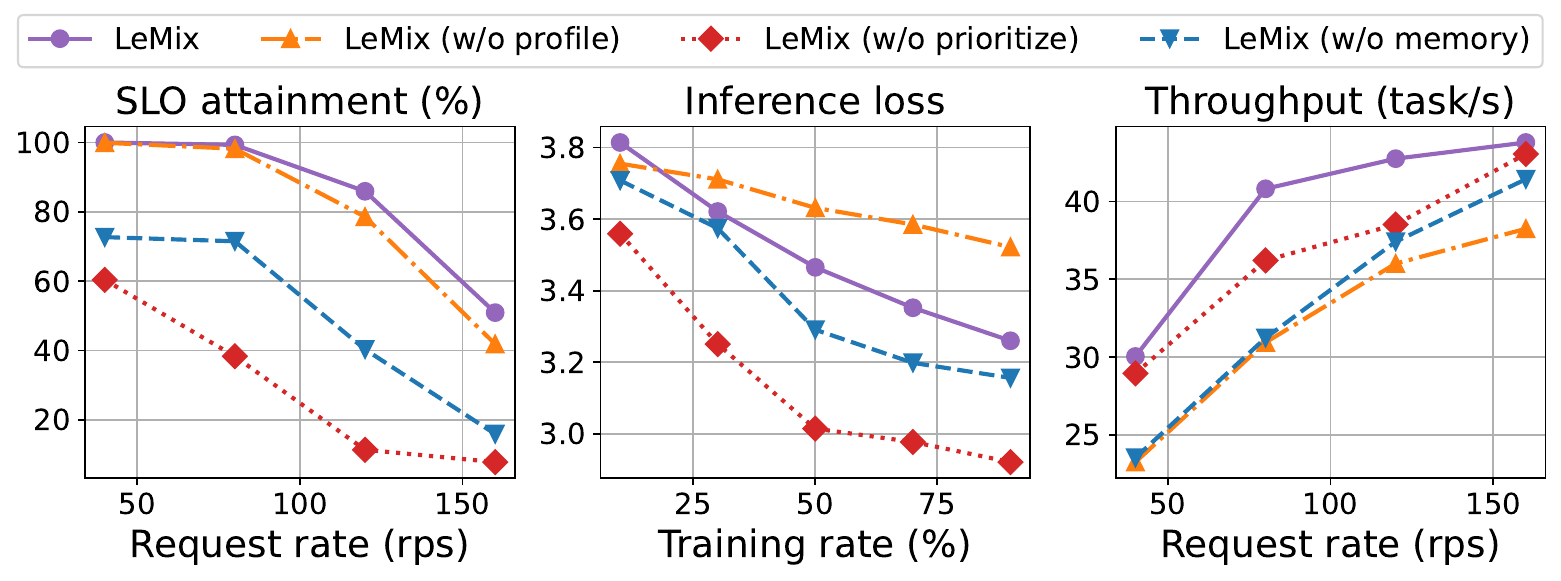}
    \caption{Ablation analysis of \approach's components.
    }
    \label{fig:ablation_analysis}
    \vspace{-5pt}
\end{figure}

\noindent\textbf{Robustness to offline priors.}
\approach employs offline profiling (\S\ref{sec:offline_profiling}) to improve the planning precision of execution latencies and memory demands, which enhances resource utilization. As shown in Figure~\ref{fig:ablation_analysis}, removing offline profiling (w/o profile) causes notable decrease in throughput, dropping by up to 0.82$\times$ compared to the full \approach design, particularly under high request rates when misjudgments tend to be accumulated. Inference loss also increases from 3.3 to 3.6 at 75\% training rate due to potential update delays for inference tasks.
Despite these setbacks, \approach (w/o profile) still outperforms baseline methods in higher throughput and SLO attainment due to fine-grained scheduling.
This robustness to profiling accuracy highlights \approach's applicability even when offline resources are limited.

\noindent\textbf{Prioritization in responsiveness.} 
Queue-level task prioritization (\S\ref{sec:task_allocation}) in \approach is essential for meeting SLOs under high request rates. 
As shown in Figure~\ref{fig:ablation_analysis}, removing prioritization (w/o prioritize) causes SLO attainment to collapse drastically from 95\% to below 30\% at 100 rps and further plummeting to 0\% at 150 rps. This is caused by the uncontrolled interference from training tasks which delay inference requests. 
\approach, in contrast, dynamically deprioritizes training tasks that risk SLO breaches, maintaining SLO attainment above 50\% and improving throughput across the workload spectrum.
Additionally, its slight drop in loss suggests the strength of dynamic scheduling where deprioritization sacrifices short-term accuracy improvements to ensure responsiveness. 

\noindent\textbf{Memory awareness in efficiency.} 
Memory-aware scheduling (\S\ref{sec:execution_scheduling}) enables \approach to adaptively manage GPU resources and prevent memory overruns, ensuring system responsiveness. When memory awareness is disabled (w/o memory), SLO attainment degrades significantly, dropping from 95\% to 55\% at 100 rps, as shown in Figure~\ref{fig:ablation_analysis}. Consequently, throughput drops from 8 to 6 at 80 rps due to memory-induced latency. By proactively offloading activations and enforcing memory thresholds, the full \approach ensures stable performance across diverse workloads while balancing resource demands.

\begin{table}[]
    \centering
    \resizebox{0.485\textwidth}{!}{
    \begin{tabular}{lcccccc}
    \hline
       & \multirow{2}{*}{Separate} & \multirow{2}{*}{RR} & \multirow{2}{*}{LUF} & \multicolumn{3}{c}{{\approach}} \\
    \cline{5-7}
        & & & & EP & RA & MS \\
    \hline
      Latency (ms) & 1.4e-2 & 1.5e-2 & 76  & 1.4e-1 & 1.4e-2 & 5.6e-2\\
    \hline
      Memory (MB) & 1.6 & 1.8 & 727.5 &  9.6 & 1.9 & 1.8 \\
    \hline
    \end{tabular}
    }
    \caption{Average time overhead (ms) and memory (MB) of schedulers. EP, RA, MS represents execution planning, resource allocation, and memory-aware scheduling.
    } 
    \label{tab:profiling}
    \vspace{-10pt}
\end{table}

\noindent\textbf{Overhead analysis.}
Table~\ref{tab:profiling} shows the average latency and memory usage under a request rate of 50 rps and training rate of 50\%.
\approach achieves its scheduling improvements with minimal computational cost. Specifically, execution planning incurs an average latency of 0.14 ms, slightly higher than simpler policies like RR, but still negligible compared to the model forward latency (Table~\ref{tab:models}). Meanwhile, resource allocation and scheduling achieve latency comparable to RR, ensuring real-time responsiveness.
Memory usage remains similarly efficient. While methods like LUF incur high overhead (727.5 MB) due to system-level GPU utilization tracking, \approach remains lightweight, consuming only 9.6 MB for its execution prediction and less than 4 MB for other modules. 
This efficient nature
of \approach ensures scalability under heavy traffic conditions and supports real-time decision-making without introducing performance bottlenecks.

\begin{figure}[t]
    \centering
    \includegraphics[width=0.485\textwidth]{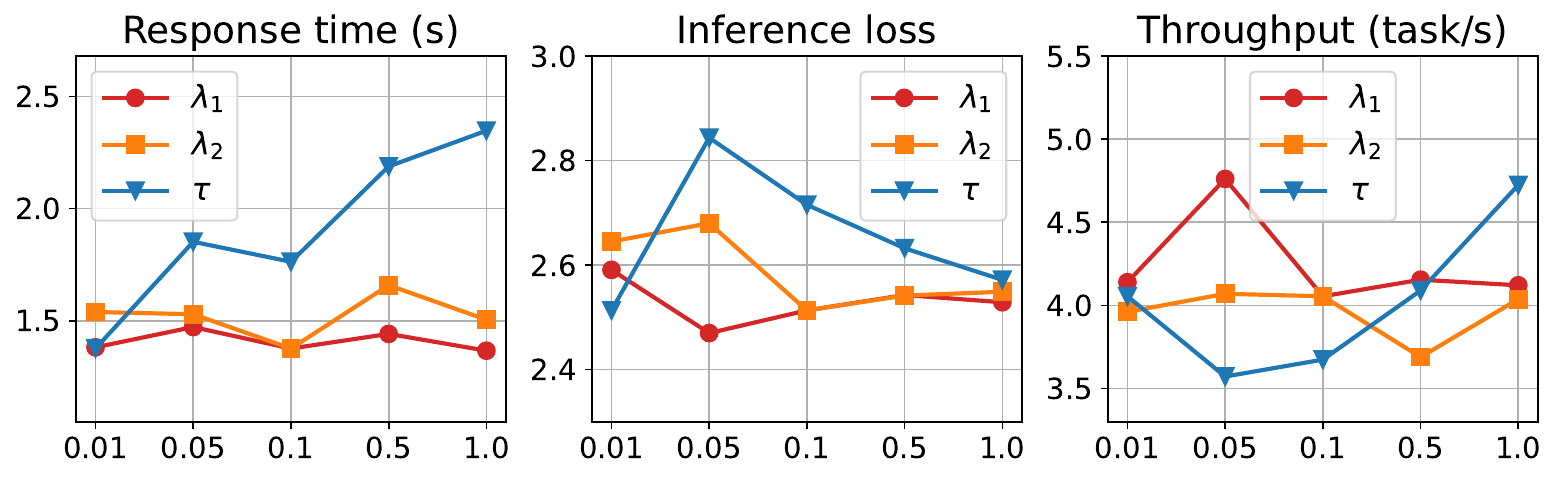}
    \caption{Trade-offs in \approach's multi-objective allocation.}
    \label{fig:parameter_study}
    \vspace{-5pt}
\end{figure}

\subsection{Parameter Study}
Figure~\ref{fig:parameter_study} illustrates the trade-offs between response time, serving quality (inference loss), and throughput under varying priority weights ($\lambda_1$, $\lambda_2$) and idleness tolerance threshold ($\tau$) in \approach's resource allocation. 
Increasing $\lambda_1$ (response time weight) reduces latency by prioritizing faster execution but limits resource utilization, while higher $\lambda_2$ (length consistency weight) improves serving quality by reducing workload heterogeneity at the cost of slight delays. Meanwhile, larger $\tau$ allows better idle-period utilization and higher throughput by consolidating workloads to fewer nodes, but risks prolonged response times as tasks wait for resources. Overall, \approach achieves a robust balance with ``\emph{sweet spots}'' of $\lambda_1$, $\lambda_2$, and $\tau$: it maintains low response times (e.g., 150 ms), high throughput (e.g., 4.5 task/s), and stable inference quality (e.g., 2.5), showcasing its adaptability to dynamic real-time conditions.


\section{Related Work}

\noindent\textbf{Distributed training.}
Distributed training of large models has been extensively explored to address the computational demands of increasing complexity. Techniques such as data parallelism~\cite{sergeev2018horovod,jiang2020unified,dean2012large} where the data is split across multiple processors, model parallelism~\cite{shoeybi2019megatron} where the model is horizontally sharded into multiple computational units, and pipeline parallelism~\cite{huang2019gpipe,li2021chimera,athlur2022varuna,osawa2023pipefisher} which divides each input mini-batch into small micro-batches to further reduce idleness, are well-established. 
While these approaches optimize utilization, they leave inter-pipeline idle periods unaddressed. 


\noindent\textbf{Inference serving.}
LLM serving systems have evolved to tackle the dynamic requirements of LLM workloads. General-purpose systems~\cite{torchserve2023,nvidia2019triton,li2023white,chen2023dycl,chen2022learning} are widely used for production environments, while LLM-optimized systems\cite{narayanan2021efficient,agrawal2024taming,li2024integrated} have emerged to handle KV cache management and chunked prefill processing. 
Techniques like Orca~\cite{yu2022orca} and FastServe~\cite{wu2023fast} introduce continuous batching and preemptive scheduling~\cite{sun2024llumnix}, addressing delays caused by long jobs, but often exacerbate interference when workloads are co-located. Furthermore, disaggregation-based approaches~\cite{patel2024splitwise, stratidejavu, zhong2024distserve} reduce contention by isolating specific workload stages. Unlike these methods, \approach focuses on node allocation, idleness utilization, and retraining alignment, demonstrating compatibility with autoregressive generation.



\noindent\textbf{Data drift and continual learning.}
Adapting to data drift~\cite{li2021estimating,li2024uncertainty} has been widely studied in continuous learning scenarios, where models evolve with new incoming data. Traditional methods include transfer learning~\cite{sun2020test, hinton2015distilling,li2025dr} and catastrophic forgetting mitigation~\cite{goodfellow2013empirical,li2023uncertainty,li2024distantly}, while edge deployment systems~\cite{bhardwaj2022ekya,shubha2023adainf,li2023lyra} demonstrate scheduling techniques for model retraining. Inference serving for LLMs also faces challenges similar to data drift, as models require frequent alignment to updated user interactions and factual knowledge. \approach builds upon continuous learning principles by co-locating training and inference workloads, enabling real-time adaptation with reduced inter-node synchronization delays, which is critical for emerging ``learning while serving'' paradigms.


\section{Conclusion}

In this work, we introduced \approach, a framework that schedules concurrent LLM training and inference workloads in distributed systems. The key innovation is exploring optimizations when co-locating both computational workloads under dynamic traffic and execution heterogeneity, and integrating these insights to optimize resource utilization while attaining SLOs.
By real-time quantifying task-specific contributions to idle periods, accuracy, and response time, \approach achieves up to 3.53$\times$ higher throughput, 0.61$\times$ lower loss, and 2.12$\times$ higher SLO attainment over traditional \separate setups.

\section*{Acknowledgment}

This research was supported by the National Science Foundation under NSF CNS 2300525, CNS 2343653, CNS 2312397.

\bibliographystyle{IEEEtran}
\bibliography{IEEEabrv,ref}

\begin{thebibliography}{10}
\providecommand{\url}[1]{#1}
\csname url@samestyle\endcsname
\providecommand{\newblock}{\relax}
\providecommand{\bibinfo}[2]{#2}
\providecommand{\BIBentrySTDinterwordspacing}{\spaceskip=0pt\relax}
\providecommand{\BIBentryALTinterwordstretchfactor}{4}
\providecommand{\BIBentryALTinterwordspacing}{\spaceskip=\fontdimen2\font plus
\BIBentryALTinterwordstretchfactor\fontdimen3\font minus \fontdimen4\font\relax}
\providecommand{\BIBforeignlanguage}[2]{{%
\expandafter\ifx\csname l@#1\endcsname\relax
\typeout{** WARNING: IEEEtran.bst: No hyphenation pattern has been}%
\typeout{** loaded for the language `#1'. Using the pattern for}%
\typeout{** the default language instead.}%
\else
\language=\csname l@#1\endcsname
\fi
#2}}
\providecommand{\BIBdecl}{\relax}
\BIBdecl

\bibitem{wei2022chain}
J.~Wei, X.~Wang, D.~Schuurmans, M.~Bosma, F.~Xia, E.~Chi, Q.~V. Le, D.~Zhou \emph{et~al.}, ``Chain-of-thought prompting elicits reasoning in large language models,'' \emph{Advances in neural information processing systems}, vol.~35, pp. 24\,824--24\,837, 2022.

\bibitem{yao2024tree}
S.~Yao, D.~Yu, J.~Zhao, I.~Shafran, T.~Griffiths, Y.~Cao, and K.~Narasimhan, ``Tree of thoughts: Deliberate problem solving with large language models,'' \emph{Advances in Neural Information Processing Systems}, vol.~36, 2024.

\bibitem{zheng2024judging}
L.~Zheng, W.-L. Chiang, Y.~Sheng, S.~Zhuang, Z.~Wu, Y.~Zhuang, Z.~Lin, Z.~Li, D.~Li, E.~Xing \emph{et~al.}, ``Judging llm-as-a-judge with mt-bench and chatbot arena,'' \emph{Advances in Neural Information Processing Systems}, vol.~36, 2024.

\bibitem{sun2020test}
Y.~Sun, X.~Wang, Z.~Liu, J.~Miller, A.~Efros, and M.~Hardt, ``Test-time training with self-supervision for generalization under distribution shifts,'' in \emph{International conference on machine learning}.\hskip 1em plus 0.5em minus 0.4em\relax PMLR, 2020, pp. 9229--9248.

\bibitem{akyürek2024surprising}
E.~Akyürek, M.~Damani, L.~Qiu, H.~Guo, Y.~Kim, and J.~Andreas, ``The surprising effectiveness of test-time training for abstract reasoning,'' \emph{arXiv preprint arXiv:2411.07279}, 2024.

\bibitem{langton2018machine}
\BIBentryALTinterwordspacing
Langton, ``Machine learning model training over time,'' May 2018. [Online]. Available: \url{https://www.langton.cloud/machine-learning-model-training-over-time/}
\BIBentrySTDinterwordspacing

\bibitem{choi2021multi}
S.~Choi, S.~Lee, Y.~Kim, J.~Park, Y.~Kwon, and J.~Huh, ``Multi-model machine learning inference serving with gpu spatial partitioning,'' \emph{arXiv preprint arXiv:2109.01611}, 2021.

\bibitem{microsoft2023optimized}
\BIBentryALTinterwordspacing
A.~Vilcek, ``Optimized training and inference of hugging face models on azure,'' September 2022. [Online]. Available: \url{https://techcommunity.microsoft.com/t5/microsoft-developer-community/optimized-training-and-inference-of-hugging-face-models-on-azure/ba-p/3631401}
\BIBentrySTDinterwordspacing

\bibitem{AmazonSageMakerTraining}
{Amazon Web Services, Inc.}, ``Train a model with amazon sagemaker,'' \url{https://docs.aws.amazon.com/sagemaker/latest/dg/how-it-works-training.html}, Amazon Web Services, Inc., 2024, accessed: 2024-05-12.

\bibitem{choi2022serving}
S.~Choi, S.~Lee, Y.~Kim, J.~Park, Y.~Kwon, and J.~Huh, ``Serving heterogeneous machine learning models on {Multi-GPU} servers with {Spatio-Temporal} sharing,'' in \emph{2022 USENIX Annual Technical Conference (USENIX ATC 22)}.\hskip 1em plus 0.5em minus 0.4em\relax Carlsbad, CA: USENIX Association, Jul. 2022, pp. 199--216.

\bibitem{kwon2023efficient}
W.~Kwon, Z.~Li, S.~Zhuang, Y.~Sheng, L.~Zheng, C.~H. Yu, J.~Gonzalez, H.~Zhang, and I.~Stoica, ``Efficient memory management for large language model serving with pagedattention,'' in \emph{Proceedings of the 29th Symposium on Operating Systems Principles}, 2023, pp. 611--626.

\bibitem{sun2024llumnix}
B.~Sun, Z.~Huang, H.~Zhao, W.~Xiao, X.~Zhang, Y.~Li, and W.~Lin, ``Llumnix: Dynamic scheduling for large language model serving,'' in \emph{18th USENIX Symposium on Operating Systems Design and Implementation (OSDI 24)}, 2024.

\bibitem{zhong2024distserve}
Y.~Zhong, S.~Liu, J.~Chen, J.~Hu, Y.~Zhu, X.~Liu, X.~Jin, and H.~Zhang, ``$\{$DistServe$\}$: Disaggregating prefill and decoding for goodput-optimized large language model serving,'' in \emph{18th USENIX Symposium on Operating Systems Design and Implementation (OSDI 24)}, 2024, pp. 193--210.

\bibitem{agrawal2024taming}
A.~Agrawal, N.~Kedia, A.~Panwar, J.~Mohan, N.~Kwatra, B.~Gulavani, A.~Tumanov, and R.~Ramjee, ``Taming $\{$Throughput-Latency$\}$ tradeoff in $\{$LLM$\}$ inference with $\{$Sarathi-Serve$\}$,'' in \emph{18th USENIX Symposium on Operating Systems Design and Implementation (OSDI 24)}, 2024, pp. 117--134.

\bibitem{wu2024dlora}
B.~Wu, R.~Zhu, Z.~Zhang, P.~Sun, X.~Liu, and X.~Jin, ``$\{$dLoRA$\}$: Dynamically orchestrating requests and adapters for $\{$LoRA$\}$$\{$LLM$\}$ serving,'' in \emph{18th USENIX Symposium on Operating Systems Design and Implementation (OSDI 24)}, 2024, pp. 911--927.

\bibitem{fu2024serverlessllm}
Y.~Fu, L.~Xue, Y.~Huang, A.-O. Brabete, D.~Ustiugov, Y.~Patel, and L.~Mai, ``$\{$ServerlessLLM$\}$:$\{$Low-Latency$\}$ serverless inference for large language models,'' in \emph{18th USENIX Symposium on Operating Systems Design and Implementation (OSDI 24)}, 2024, pp. 135--153.

\bibitem{yu2022orca}
G.-I. Yu, J.~S. Jeong, G.-W. Kim, S.~Kim, and B.-G. Chun, ``Orca: A distributed serving system for {Transformer-Based} generative models,'' in \emph{16th USENIX Symposium on Operating Systems Design and Implementation (OSDI 22)}.\hskip 1em plus 0.5em minus 0.4em\relax Carlsbad, CA: USENIX Association, Jul. 2022, pp. 521--538.

\bibitem{gaunt2017ampnet}
A.~L. Gaunt, M.~A. Johnson, M.~Riechert, D.~Tarlow, R.~Tomioka, D.~Vytiniotis, and S.~Webster, ``Ampnet: Asynchronous model-parallel training for dynamic neural networks,'' \emph{arXiv preprint arXiv:1705.09786}, 2017.

\bibitem{narayanan2019pipedream}
D.~Narayanan, A.~Harlap, A.~Phanishayee, V.~Seshadri, N.~R. Devanur, G.~R. Ganger, P.~B. Gibbons, and M.~Zaharia, ``Pipedream: generalized pipeline parallelism for dnn training,'' in \emph{Proceedings of the 27th ACM symposium on operating systems principles}, 2019, pp. 1--15.

\bibitem{shoeybi2019megatron}
M.~Shoeybi, M.~Patwary, R.~Puri, P.~LeGresley, J.~Casper, and B.~Catanzaro, ``Megatron-lm: Training multi-billion parameter language models using model parallelism,'' \emph{arXiv preprint arXiv:1909.08053}, 2019.

\bibitem{athlur2022varuna}
S.~Athlur, N.~Saran, M.~Sivathanu, R.~Ramjee, and N.~Kwatra, ``Varuna: scalable, low-cost training of massive deep learning models,'' in \emph{Proceedings of the Seventeenth European Conference on Computer Systems}, 2022, pp. 472--487.

\bibitem{choi2023envpipe}
S.~Choi, I.~Koo, J.~Ahn, M.~Jeon, and Y.~Kwon, ``$\{$EnvPipe$\}$: Performance-preserving $\{$DNN$\}$ training framework for saving energy,'' in \emph{2023 USENIX Annual Technical Conference (USENIX ATC 23)}, 2023, pp. 851--864.

\bibitem{li2023alpaserve}
Z.~Li, L.~Zheng, Y.~Zhong, V.~Liu, Y.~Sheng, X.~Jin, Y.~Huang, Z.~Chen, H.~Zhang, J.~E. Gonzalez \emph{et~al.}, ``$\{$AlpaServe$\}$: Statistical multiplexing with model parallelism for deep learning serving,'' in \emph{17th USENIX Symposium on Operating Systems Design and Implementation (OSDI 23)}, 2023, pp. 663--679.

\bibitem{lai2023merak}
Z.~Lai, S.~Li, X.~Tang, K.~Ge, W.~Liu, Y.~Duan, L.~Qiao, and D.~Li, ``Merak: An efficient distributed dnn training framework with automated 3d parallelism for giant foundation models,'' \emph{IEEE Transactions on Parallel and Distributed Systems}, vol.~34, no.~5, pp. 1466--1478, 2023.

\bibitem{huang2019gpipe}
Y.~Huang, Y.~Cheng, A.~Bapna, O.~Firat, D.~Chen, M.~Chen, H.~Lee, J.~Ngiam, Q.~V. Le, Y.~Wu \emph{et~al.}, ``Gpipe: Efficient training of giant neural networks using pipeline parallelism,'' \emph{Advances in neural information processing systems}, vol.~32, 2019.

\bibitem{touvron2023llama}
H.~Touvron, L.~Martin, K.~Stone, P.~Albert, A.~Almahairi, Y.~Babaei, N.~Bashlykov, S.~Batra, P.~Bhargava, S.~Bhosale \emph{et~al.}, ``Llama 2: Open foundation and fine-tuned chat models,'' \emph{arXiv preprint arXiv:2307.09288}, 2023.

\bibitem{zhenglmsys}
L.~Zheng, W.-L. Chiang, Y.~Sheng, T.~Li, S.~Zhuang, Z.~Wu, Y.~Zhuang, Z.~Li, Z.~Lin, E.~Xing \emph{et~al.}, ``Lmsys-chat-1m: A large-scale real-world llm conversation dataset,'' in \emph{The Twelfth International Conference on Learning Representations}, 2023.

\bibitem{zheng2023judging}
L.~Zheng, W.-L. Chiang, Y.~Sheng, S.~Zhuang, Z.~Wu, Y.~Zhuang, Z.~Lin, Z.~Li, D.~Li, E.~Xing \emph{et~al.}, ``Judging llm-as-a-judge with mt-bench and chatbot arena,'' \emph{Advances in Neural Information Processing Systems}, vol.~36, pp. 46\,595--46\,623, 2023.

\bibitem{sheng2024fairness}
Y.~Sheng, S.~Cao, D.~Li, B.~Zhu, Z.~Li, D.~Zhuo, J.~E. Gonzalez, and I.~Stoica, ``Fairness in serving large language models,'' in \emph{18th USENIX Symposium on Operating Systems Design and Implementation (OSDI 24)}, 2024, pp. 965--988.

\bibitem{shen2024large}
Y.~Shen, J.~Shao, X.~Zhang, Z.~Lin, H.~Pan, D.~Li, J.~Zhang, and K.~B. Letaief, ``Large language models empowered autonomous edge ai for connected intelligence,'' \emph{IEEE Communications Magazine}, 2024.

\bibitem{christiano2017deep}
P.~F. Christiano, J.~Leike, T.~Brown, M.~Martic, S.~Legg, and D.~Amodei, ``Deep reinforcement learning from human preferences,'' \emph{Advances in neural information processing systems}, vol.~30, 2017.

\bibitem{li2025dr}
Y.~Li, J.~Nham, G.~Jawahar, L.~Shu, D.~Uthus, Y.-H. Sung, C.~Yang, I.~Rolnick, Y.~Qiao, and C.~Liu, ``Dr genre: Reinforcement learning from decoupled llm feedback for generic text rewriting,'' \emph{arXiv preprint arXiv:2503.06781}, 2025.

\bibitem{dai2023safe}
J.~Dai, X.~Pan, R.~Sun, J.~Ji, X.~Xu, M.~Liu, Y.~Wang, and Y.~Yang, ``Safe rlhf: Safe reinforcement learning from human feedback,'' in \emph{The Twelfth International Conference on Learning Representations}, 2023.

\bibitem{fu2024safety}
Y.~Fu, Y.~Li, W.~Xiao, C.~Liu, and Y.~Dong, ``Safety alignment in nlp tasks: Weakly aligned summarization as an in-context attack,'' in \emph{Proceedings of the 62nd Annual Meeting of the Association for Computational Linguistics (Volume 1: Long Papers)}, 2024, pp. 8483--8502.

\bibitem{jia2019beyond}
Z.~Jia, M.~Zaharia, and A.~Aiken, ``Beyond data and model parallelism for deep neural networks.'' \emph{Proceedings of Machine Learning and Systems}, vol.~1, pp. 1--13, 2019.

\bibitem{gptai2024servers}
\BIBentryALTinterwordspacing
{GPT AI Team}, ``How many servers are needed to run chatgpt?'' Aug. 2024, accessed: 2024-12-05. [Online]. Available: \url{https://gptai.tn/how-many-servers-to-run-chatgpt/}
\BIBentrySTDinterwordspacing

\bibitem{bai2022training}
Y.~Bai, A.~Jones, K.~Ndousse, A.~Askell, A.~Chen, N.~DasSarma, D.~Drain, S.~Fort, D.~Ganguli, T.~Henighan \emph{et~al.}, ``Training a helpful and harmless assistant with reinforcement learning from human feedback,'' \emph{arXiv preprint arXiv:2204.05862}, 2022.

\bibitem{ethayarajh22a}
K.~Ethayarajh, Y.~Choi, and S.~Swayamdipta, ``Understanding dataset difficulty with $\mathcal{V}$-usable information,'' in \emph{Proceedings of the 39th International Conference on Machine Learning}, ser. Proceedings of Machine Learning Research, vol. 162.\hskip 1em plus 0.5em minus 0.4em\relax PMLR, 17--23 Jul 2022, pp. 5988--6008.

\bibitem{vaswani2017attention}
A.~Vaswani, N.~Shazeer, N.~Parmar, J.~Uszkoreit, L.~Jones, A.~N. Gomez, {\L}.~Kaiser, and I.~Polosukhin, ``Attention is all you need,'' \emph{Advances in neural information processing systems}, vol.~30, 2017.

\bibitem{wu2023transparent}
B.~Wu, Z.~Zhang, Z.~Bai, X.~Liu, and X.~Jin, ``Transparent $\{$GPU$\}$ sharing in container clouds for deep learning workloads,'' in \emph{20th USENIX Symposium on Networked Systems Design and Implementation (NSDI 23)}, 2023, pp. 69--85.

\bibitem{pope2023efficiently}
R.~Pope, S.~Douglas, A.~Chowdhery, J.~Devlin, J.~Bradbury, J.~Heek, K.~Xiao, S.~Agrawal, and J.~Dean, ``Efficiently scaling transformer inference,'' \emph{Proceedings of Machine Learning and Systems}, vol.~5, pp. 606--624, 2023.

\bibitem{reagen2017methods}
B.~Reagen, Y.~S. Shao, S.~L. Xi, G.-Y. Wei, and D.~Brooks, ``Methods and infrastructure in the era of accelerator-centric architectures,'' in \emph{2017 IEEE 60th International Midwest Symposium on Circuits and Systems (MWSCAS)}.\hskip 1em plus 0.5em minus 0.4em\relax IEEE, 2017, pp. 902--905.

\bibitem{li2023rt}
Y.~Li, Z.~Li, W.~Yang, and C.~Liu, ``Rt-lm: Uncertainty-aware resource management for real-time inference of language models,'' in \emph{2023 IEEE Real-Time Systems Symposium (RTSS)}.\hskip 1em plus 0.5em minus 0.4em\relax IEEE, 2023, pp. 158--171.

\bibitem{li2025mixtraining}
Z.~Li, J.~Zhang, Y.~Li, Y.~Zhu, and C.~Liu, ``Mixtraining: A better trade-off between compute and performance,'' \emph{arXiv preprint arXiv:2502.19513}, 2025.

\bibitem{zheng2024sglang}
L.~Zheng, L.~Yin, Z.~Xie, C.~L. Sun, J.~Huang, C.~H. Yu, S.~Cao, C.~Kozyrakis, I.~Stoica, J.~E. Gonzalez \emph{et~al.}, ``Sglang: Efficient execution of structured language model programs,'' \emph{Advances in neural information processing systems}, vol.~37, pp. 62\,557--62\,583, 2024.

\bibitem{eisenman2022check}
A.~Eisenman, K.~K. Matam, S.~Ingram, D.~Mudigere, R.~Krishnamoorthi, K.~Nair, M.~Smelyanskiy, and M.~Annavaram, ``$\{$Check-N-Run$\}$: A checkpointing system for training deep learning recommendation models,'' in \emph{19th USENIX Symposium on Networked Systems Design and Implementation (NSDI 22)}, 2022, pp. 929--943.

\bibitem{aminabadi2022deepspeed}
R.~Y. Aminabadi, S.~Rajbhandari, A.~A. Awan, C.~Li, D.~Li, E.~Zheng, O.~Ruwase, S.~Smith, M.~Zhang, J.~Rasley \emph{et~al.}, ``Deepspeed-inference: enabling efficient inference of transformer models at unprecedented scale,'' in \emph{SC22: International Conference for High Performance Computing, Networking, Storage and Analysis}.\hskip 1em plus 0.5em minus 0.4em\relax IEEE, 2022, pp. 1--15.

\bibitem{dao2022flashattention}
T.~Dao, D.~Fu, S.~Ermon, A.~Rudra, and C.~R{\'e}, ``Flashattention: Fast and memory-efficient exact attention with io-awareness,'' \emph{Advances in neural information processing systems}, vol.~35, pp. 16\,344--16\,359, 2022.

\bibitem{zhang2020dialogpt}
Y.~Zhang, S.~Sun, M.~Galley, Y.-C. Chen, C.~Brockett, X.~Gao, J.~Gao, J.~Liu, and W.~B. Dolan, ``Dialogpt: Large-scale generative pre-training for conversational response generation,'' in \emph{Proceedings of the 58th Annual Meeting of the Association for Computational Linguistics: System Demonstrations}, 2020, pp. 270--278.

\bibitem{feizabadi2003utilitiy}
S.~Feizabadi, W.~Beebee~Jr, B.~Ravindran, P.~Li, and M.~Rinard, ``Utilitiy accrual scheduling with real-time java,'' in \emph{OTM Confederated International Conferences" On the Move to Meaningful Internet Systems"}.\hskip 1em plus 0.5em minus 0.4em\relax Springer, 2003, pp. 550--563.

\bibitem{sergeev2018horovod}
A.~Sergeev and M.~Del~Balso, ``Horovod: fast and easy distributed deep learning in tensorflow,'' \emph{arXiv preprint arXiv:1802.05799}, 2018.

\bibitem{jiang2020unified}
Y.~Jiang, Y.~Zhu, C.~Lan, B.~Yi, Y.~Cui, and C.~Guo, ``A unified architecture for accelerating distributed $\{$DNN$\}$ training in heterogeneous $\{$GPU/CPU$\}$ clusters,'' in \emph{14th USENIX Symposium on Operating Systems Design and Implementation (OSDI 20)}, 2020, pp. 463--479.

\bibitem{dean2012large}
J.~Dean, G.~Corrado, R.~Monga, K.~Chen, M.~Devin, M.~Mao, M.~Ranzato, A.~Senior, P.~Tucker, K.~Yang \emph{et~al.}, ``Large scale distributed deep networks,'' \emph{Advances in neural information processing systems}, vol.~25, 2012.

\bibitem{li2021chimera}
S.~Li and T.~Hoefler, ``Chimera: efficiently training large-scale neural networks with bidirectional pipelines,'' in \emph{Proceedings of the International Conference for High Performance Computing, Networking, Storage and Analysis}, 2021, pp. 1--14.

\bibitem{osawa2023pipefisher}
K.~Osawa, S.~Li, and T.~Hoefler, ``Pipefisher: Efficient training of large language models using pipelining and fisher information matrices,'' \emph{Proceedings of Machine Learning and Systems}, vol.~5, 2023.

\bibitem{torchserve2023}
L.~Ning, H.~Shojanazeri, K.~Wen, and the PyTorch~Foundation, ``Torchserve: Serve, optimize and scale pytorch models in production,'' PyTorch Foundation, 2023, \url{https://pytorch.org/serve/}.

\bibitem{nvidia2019triton}
{NVIDIA Corporation}, ``Triton inference server: An optimized cloud and edge inferencing solution,'' 2019, \url{https://developer.nvidia.com/nvidia-triton-inference-server}.

\bibitem{li2023white}
Y.~Li, Z.~Li, Y.~Gao, and C.~Liu, ``White-box multi-objective adversarial attack on dialogue generation,'' in \emph{Proceedings of the 61st Annual Meeting of the Association for Computational Linguistics (Volume 1: Long Papers)}, 2023, pp. 1778--1792.

\bibitem{chen2023dycl}
S.~Chen, S.~Wei, C.~Liu, and W.~Yang, ``Dycl: Dynamic neural network compilation via program rewriting and graph optimization,'' in \emph{Proceedings of the 32nd ACM SIGSOFT International Symposium on Software Testing and Analysis}, 2023, pp. 614--626.

\bibitem{chen2022learning}
S.~Chen, H.~Khanpour, C.~Liu, and W.~Yang, ``Learning to reverse dnns from ai programs automatically,'' \emph{arXiv preprint arXiv:2205.10364}, 2022.

\bibitem{narayanan2021efficient}
D.~Narayanan, M.~Shoeybi, J.~Casper, P.~LeGresley, M.~Patwary, V.~Korthikanti, D.~Vainbrand, P.~Kashinkunti, J.~Bernauer, B.~Catanzaro \emph{et~al.}, ``Efficient large-scale language model training on gpu clusters using megatron-lm,'' in \emph{Proceedings of the International Conference for High Performance Computing, Networking, Storage and Analysis}, 2021, pp. 1--15.

\bibitem{li2024integrated}
X.~Li, Y.~Ma, Y.~Huang, X.~Wang, Y.~Lin, and C.~Zhang, ``Integrated optimization of large language models: Synergizing data utilization and compression techniques,'' 2024.

\bibitem{wu2023fast}
B.~Wu, Y.~Zhong, Z.~Zhang, S.~Liu, F.~Liu, Y.~Sun, G.~Huang, X.~Liu, and X.~Jin, ``Fast distributed inference serving for large language models,'' \emph{arXiv preprint arXiv:2305.05920}, 2023.

\bibitem{patel2024splitwise}
P.~Patel, E.~Choukse, C.~Zhang, A.~Shah, {\'I}.~Goiri, S.~Maleki, and R.~Bianchini, ``Splitwise: Efficient generative llm inference using phase splitting,'' in \emph{2024 ACM/IEEE 51st Annual International Symposium on Computer Architecture (ISCA)}.\hskip 1em plus 0.5em minus 0.4em\relax IEEE, 2024, pp. 118--132.

\bibitem{stratidejavu}
F.~Strati, S.~Mcallister, A.~Phanishayee, J.~Tarnawski, and A.~Klimovic, ``Déjàvu: {KV}-cache streaming for fast, fault-tolerant generative {LLM} serving,'' in \emph{Proceedings of the 41st International Conference on Machine Learning}, ser. Proceedings of Machine Learning Research, R.~Salakhutdinov, Z.~Kolter, K.~Heller, A.~Weller, N.~Oliver, J.~Scarlett, and F.~Berkenkamp, Eds., vol. 235.\hskip 1em plus 0.5em minus 0.4em\relax PMLR, 21--27 Jul 2024, pp. 46\,745--46\,771.

\bibitem{li2021estimating}
Y.~Li, S.~Chen, and W.~Yang, ``Estimating predictive uncertainty under program data distribution shift,'' \emph{arXiv preprint arXiv:2107.10989}, 2021.

\bibitem{li2024uncertainty}
Y.~Li, S.~Chen, Y.~Guo, W.~Yang, Y.~Dong, and C.~Liu, ``Uncertainty awareness of large language models under code distribution shifts: A benchmark study,'' \emph{arXiv preprint arXiv:2402.05939}, 2024.

\bibitem{hinton2015distilling}
G.~Hinton, O.~Vinyals, and J.~Dean, ``Distilling the knowledge in a neural network,'' in \emph{NeurIPS Deep Learning and Representation Learning Workshop}, 2015.

\bibitem{goodfellow2013empirical}
I.~J. Goodfellow, M.~Mirza, D.~Xiao, A.~Courville, and Y.~Bengio, ``An empirical investigation of catastrophic forgetting in gradient-based neural networks,'' \emph{arXiv preprint arXiv:1312.6211}, 2013.

\bibitem{li2023uncertainty}
Y.~Li, X.~Yu, Y.~Liu, H.~Chen, and C.~Liu, ``Uncertainty-aware bootstrap learning for joint extraction on distantly-supervised data,'' in \emph{Proceedings of the 61st Annual Meeting of the Association for Computational Linguistics (Volume 2: Short Papers)}, 2023, pp. 1349--1358.

\bibitem{li2024distantly}
Y.~Li, X.~Yu, Y.~Guo, Y.~Liu, H.~Chen, and C.~Liu, ``Distantly-supervised joint extraction with noise-robust learning,'' in \emph{Findings of the Association for Computational Linguistics ACL 2024}, 2024, pp. 10\,202--10\,217.

\bibitem{bhardwaj2022ekya}
R.~Bhardwaj, Z.~Xia, G.~Ananthanarayanan, J.~Jiang, Y.~Shu, N.~Karianakis, K.~Hsieh, P.~Bahl, and I.~Stoica, ``Ekya: Continuous learning of video analytics models on edge compute servers,'' in \emph{19th USENIX Symposium on Networked Systems Design and Implementation (NSDI 22)}, 2022, pp. 119--135.

\bibitem{shubha2023adainf}
S.~S. Shubha and H.~Shen, ``Adainf: Data drift adaptive scheduling for accurate and slo-guaranteed multiple-model inference serving at edge servers,'' in \emph{Proceedings of the ACM SIGCOMM 2023 Conference}, 2023, pp. 473--485.

\bibitem{li2023lyra}
J.~Li, H.~Xu, Y.~Zhu, Z.~Liu, C.~Guo, and C.~Wang, ``Lyra: Elastic scheduling for deep learning clusters,'' in \emph{Proceedings of the Eighteenth European Conference on Computer Systems}, 2023, pp. 835--850.

\end{thebibliography}

\end{document}